%% file: main2.tex
\documentclass{article} 

\usepackage{url}
\usepackage{graphicx}
\usepackage{subcaption}
\usepackage[T1]{fontenc}
\usepackage{algorithm}
\usepackage{algorithmicx}
\usepackage{algpseudocode}
\usepackage{booktabs}       
\usepackage{amsfonts}       
\usepackage{nicefrac}       
\usepackage{microtype}      
\usepackage{xspace}
\usepackage{color}
\usepackage{multirow}
\usepackage{subcaption}
\usepackage{wrapfig}
\usepackage{listings}
\usepackage{color}
\usepackage{mathtools}
\usepackage{semantic}
\usepackage{times}

\usepackage{hyperref}

\newif\ifsubmit
\newif\ifarxivsubmit

\arxivsubmitfalse

\ifarxivsubmit
    \usepackage[preprint]{neurips_2019}
\else
    \usepackage[final,nonatbib]{neurips_2019}
\fi

\newcommand{\appref}{the supplementary material}
\newif\ifapp
\apptrue

\newif\ifsubmit
\submittrue

\newif\ifappsubmit
\appsubmitfalse

\ifsubmit
\newcommand{\yuandong}[1]{}
\newcommand{\xinyun}[1]{}
\newcommand{\yuandongg}[1]{}
\else
\newcommand{\yuandong}[1]{\textcolor{red}{[Yuandong: #1]}}
\newcommand{\xinyun}[1]{\textcolor{blue}{[Xinyun: #1]}}
\newcommand{\yuandongg}[1]{\textcolor{red}{[Yuandong: #1]}}
\fi

\newcommand{\yuandonggresolved}[1]{}

\newcommand{\eat}[1]{}

\newcommand{\ifes}[3]{\ensuremath{\mathbf{if}~#1~\mathbf{then}~#2~\mathbf{else}~#3~\mathbf{fi}}}
\newcommand{\ifs}[2]{\ensuremath{\mathbf{if}~#1~\mathbf{then}~#2~\mathbf{fi}}}

\def\ours{\texttt{NeuRewriter}}

\ifappsubmit
\title{Supplementary Material for ``Learning to Perform Local Rewriting for Combinatorial Optimization''}
\else
\title{Learning to Perform Local Rewriting for Combinatorial Optimization}
\fi

\author{Xinyun Chen~\thanks{Work partially done when interning at Facebook AI Research.} \\ UC Berkeley \\ \texttt{xinyun.chen@berkeley.edu}
\And
Yuandong Tian \\ Facebook AI Research \\ \texttt{yuandong@fb.com}
}

\begin{document}

\maketitle

\ifappsubmit
\else
\begin{abstract}
Search-based methods for hard combinatorial optimization are often guided by heuristics. Tuning heuristics in various conditions and situations is often time-consuming. In this paper, we propose {\ours} that learns a policy to pick heuristics and rewrite the local components of the current solution to iteratively improve it until convergence. The policy factorizes into a region-picking and a rule-picking component, each parameterized by a neural network trained with actor-critic methods in reinforcement learning. {\ours} captures the general structure of combinatorial problems and shows strong performance in three versatile tasks: expression simplification, online job scheduling and vehicle routing problems. {\ours} outperforms the expression simplification component in Z3~\cite{de2008z3}; outperforms DeepRM~\cite{mao2016resource} and Google OR-tools~\cite{GoogleOrTools} in online job scheduling; and outperforms recent neural baselines~\cite{nazari2018reinforcement,kool2018attention} and Google OR-tools~\cite{GoogleOrTools} in vehicle routing problems.~\footnote{The code is available at~\url{https://github.com/facebookresearch/neural-rewriter}.} 

\end{abstract}

\input{intro}
\input{work}
\input{setup}

\input{model}
\input{exp}

\input{conc}


{
\bibliographystyle{abbrv}
\bibliography{ref}
}

\ifappsubmit
\else
\clearpage
\fi
\appendix
\input{app}

\end{document}

%% file: intro.tex
\section{Introduction}
\label{sec:introduction}
Solving combinatorial problems is a long-standing challenge and has a lot of practical applications (e.g., job scheduling, theorem proving, planning, decision making). While problems with specific structures (e.g., shortest path) can be solved efficiently with proven algorithms (e.g, dynamic programming, greedy approach, search), many combinatorial problems are NP-hard and rely on manually designed heuristics to improve the quality of solutions~\cite{affenzeller2002generic,reeves1995modern,karp1972reducibility}.

Although it is usually easy to come up with many heuristics, determining when and where such heuristics should be applied, and how they should be prioritized, is time-consuming. It takes commercial solvers decades to tune to strong performance in practical problems~\cite{de2008z3,sorensson2005minisat,GoogleOrTools}. 

To address this issue, previous works use neural networks to predict a complete solution from scratch, given a complete description of the problem~\cite{vinyals2015pointer,mao2016resource,kool2018attention,graves2014neural}. While this avoids search and tuning, a direct prediction could be difficult when the number of variables grows.

Improving iteratively from an existing solution is a common approach for continuous solution spaces, e.g, trajectory optimization in robotics~\cite{mayne1973differential,tassa2012synthesis,levine2014learning}. However, such methods relying on gradient information to guide the search, is not applicable for discrete solution spaces due to indifferentiablity. 

To address this problem, we directly learn a neural-based policy that improves the current solution by iteratively rewriting a local part of it until convergence. Inspired by the problem structures, the policy is factorized into two parts: the region-picking and the rule-picking policy, and is trained end-to-end with reinforcement learning, rewarding cumulative improvement of the solution. 

We apply our approach, {\ours}, to three different domains: expression simplification, online job scheduling, and vehicle routing problems. We show that {\ours} is better than strong heuristics using multiple metrics. For expression simplification, {\ours} outperforms the expression simplification component in Z3~\cite{de2008z3}. For online job scheduling, under a controlled setting, {\ours} outperforms Google OR-tools~\cite{GoogleOrTools} in terms of both speed and quality of the solution, and DeepRM~\cite{mao2016resource}, a neural-based approach that predicts a holistic scheduling plan, by large margins especially in more complicated setting (e.g., with more heterogeneous resources).
For vehicle routing problems, {\ours} outperforms two recent neural network approaches~\cite{nazari2018reinforcement,kool2018attention} and Google OR-tools~\cite{GoogleOrTools}.
Furthermore, extensive ablation studies show that our approach works well in different situations (e.g., different expression lengths, non-uniform job/resource distribution), and transfers well when distribution shifts (e.g., test on longer expressions than those used for training). 

%% file: work.tex
\begin{figure*}[ht]
    \centering
    \includegraphics[width=0.9\linewidth]{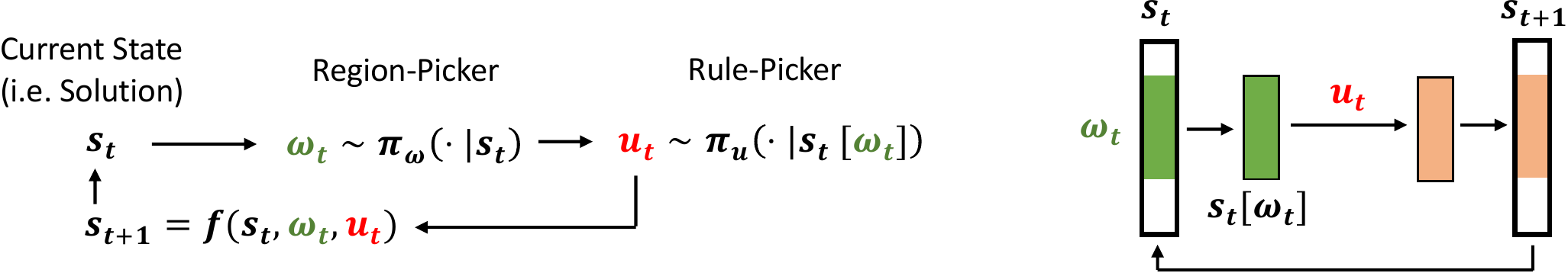}
    \caption{
    \small{The framework of our neural rewriter. Given the current state (i.e., solution to the optimization problem) $s_t$, we first pick a region $\omega_t$ by the region-picking policy $\pi_\omega(\omega_t|s_t)$, and then pick a rewriting rule $u_t$ using the rule-picking policy $\pi_u(u_t|s_t[\omega_t])$, where $\pi_u(u_t|s_t[\omega_t])$ gives the probability distribution of applying each rewriting rule $u \in \mathcal{U}$ to the partial solution. Once the partial solution is updated, we obtain an improved solution $s_{t+1}$ and repeat the process until convergence.}}
    \label{fig:rewrite-arch}
\end{figure*}

\vspace{-0.5cm}
\section{Related Work}
\label{sec:work}
\textbf{Methods}. Using neural network models for combinatorial optimization has been explored in the last few years. A straightforward idea is to construct a solution directly (e.g., with a Seq2Seq model) from the problem specification~\cite{vinyals2015pointer,bay2017approximating,mao2016resource,khalil2017learning}. However, such approaches might meet with difficulties if the problem has complex configurations, as our evaluation indicates. In contrast, our paper focuses on iterative improvement of a \emph{complete} solution. 

Trajectory optimization with local gradient information has been widely studied in robotics with many effective techniques~\cite{mayne1973differential,bradtke1994adaptive,vrabie2009adaptive,tassa2012synthesis,levine2013guided,levine2014learning}. For discrete problems, it is possible to apply continuous relaxation and apply gradient descent~\cite{bunel2016adaptive}. In contrast, we \emph{learn the gradient} from previous experience to optimize a complete solution, similar to data-driven descent~\cite{tian2013hierarchical} and synthetic gradient~\cite{jaderberg2017decoupled}. 

At a high level, our framework is closely connected with the local search pipeline. Specifically, we can leverage our learned RL policy to guide the local search, i.e., to decide which neighbor solution to move to. We will demonstrate that in our evaluated tasks, our approach outperforms several local search algorithms guided by manually designed heuristics, and softwares supporting more advanced local search algorithms, i.e., Z3~\cite{de2008z3} and OR-tools~\cite{GoogleOrTools}.

\textbf{Applications}. For expression simplification, some recent work use deep neural networks to discover equivalent expressions~\cite{cai2018learning,allamanis2017learning,zaremba2014learning}. In particular,~\cite{cai2018learning} trains a deep neural network to rewrite algebraic expressions with supervised learning, which requires a collection of ground truth rewriting paths, and may not find novel rewriting routines. We mitigate these limitations using reinforcement learning.

Job scheduling and resource management problems are ubiquitous and fundamental in computer systems. Various work have studied these problems from both theoretical and empirical sides~\cite{blazewicz1996job,grandl2015multi,armbrust2010view,scully2017optimally,terekhov2014queueing,mao2016resource,chen2017deep}. In particular, a recent line of work studies deep reinforcement learning for job scheduling~\cite{mao2016resource,chen2017deep} and vehicle routing problems~\cite{kool2018attention,nazari2018reinforcement}. 

Our approach is tested on multiple domains with extensive ablation studies, and could also be extended to other closely related tasks such as code optimization~\cite{schkufza2013stochastic,chen2018learning}, theorem proving~\cite{huang2018gamepad,lederman2018learning,bachmair1994rewrite,hsiang1992term}, text simplification~\cite{cohn2009sentence,paetzold2013text,feblowitz2013sentence}, and classical combinatorial optimization problems beyond routing problems~\cite{deudon2018learning,khalil2017learning,bello2016neural,vinyals2015pointer,karp1972reducibility}, e.g., Vertex Cover Problem~\cite{bar1981linear}.

%% file: setup.tex
\vspace{-0.3cm}
\section{Problem Setup} \label{sec:setup}
\vspace{-0.3cm}

\def\rr{\mathbb{R}}

Let $\mathcal{S}$ be the space of all feasible solutions in the problem domain, and $c:\mathcal{S}\to\rr$ be the cost function. The goal of optimization is to find $\arg\min_{s \in \mathcal{S}}c(s)$. In this work, instead of finding a solution from scratch, we first construct a feasible one, then make incremental improvement by iteratively applying \emph{local rewriting rules} to the existing solution until convergence. Our rewriting formulation is especially suitable for problems with the following properties: \textbf{(1)} a feasible solution is easy to find; \textbf{(2)} the search space has well-behaved local structures, which could be utilized to incrementally improve the solution. For such problems, a complete solution provides a full context for the improvement using a rewriting-based approach, allowing additional features to be computed, which is hard to obtain if the solution is generated from scratch; meanwhile, different solutions might share a common routine towards the optimum, which could be represented as local rewriting rules. For example, it is much easier to decide whether to postpone jobs with large resource requirements when an existing job schedule is provided. Furthermore, simple rules like swapping two jobs could improve the performance. 

Formally, each solution is a \emph{state}, and each local region and the associated rewriting rule is an \emph{action}.

\textbf{Optimization as a rewriting problem.} Let $\mathcal{U}$ be the rewriting ruleset. Suppose $s_t$ is the current solution (or state) at iteration $t$. 
We first compute a state-dependent \emph{region set} $\Omega(s_t)$, then pick a region $\omega_t\in \Omega(s_t)$ using the \emph{region-picking policy} $\pi_\omega(\omega_t | s_t)$. We then pick a rewriting rule $u_t$ applicable to that region $\omega_t$ using the \emph{rule-picking policy} $\pi_u(u_t|s_t[\omega_t])$, where $s_t[\omega_t]$ is a subset of state $s_t$. We then apply this rewriting rule $u_t \in \mathcal{U}$ to $s_t[\omega_t]$, and obtain the next state $s_{t+1}= f(s_t, \omega_t, u_t)$. Given an initial solution (or state) $s_0$, our goal is to find a sequence of rewriting steps $(s_0, (\omega_0, u_0)), (s_1, (\omega_1, u_1)), ..., (s_{T-1}, (\omega_{T-1}, u_{T-1})), s_T$ so that the final cost $c(s_T)$ is minimized.

To tackle a rewriting problem, rule-based rewriters with manually-designed rewriting routines have been proposed~\cite{halide2018simplifier}. However, manually designing such routines is not a trivial task. An incomplete set of routines often leads to an inefficient exhaustive search, while a set of kaleidoscopic routines is often cumbersome to design, hard to maintain and lacks flexibility.

In this paper, we propose to train a neural network instead, using reinforcement learning. Recent advance in deep reinforcement learning suggests the potential of well-trained models to discover novel effective policies, such as demonstrated in Computer Go~\cite{silver2017mastering} and video games~\cite{OpenAIdota}. Moreover, by leveraging reinforcement learning, our approach could be extended to a broader range of problems that could be hard for rule-based rewriters and classic search algorithms. For example, we can design the reward to take the validity of the solution into account, so that we can start with an infeasible solution and then move towards a feasible one. On the other hand, we can also train the neural network to explore the connections between different solutions in the search space. In our evaluation, we demonstrate that our approach (1) mitigates laborious human efforts, (2) discovers novel rewriting paths from its own exploration, and (3) finds better solution to optimization problem than the current state-of-the-art and traditional heuristic-based software packages tuned for decades. 

\eat{
On the other hand, an exhaustive search over all possible rewriting paths is generally implausible due to the large search space. Thus, a neural network-based rewriter learned from data has the potential of combining the best of both worlds, i.e., reducing the effort of heuristics design while not compromising the time-efficiency. In addition, training the model using reinforcement learning
}

%% file: model.tex
\vspace{-0.3cm}
\section{Neural Rewriter Model} \label{sec:approach}
\vspace{-0.3cm}

In the following, we present the design of our rewriting model, i.e., {\ours}. We first provide an overview of our model framework, then present the design details for different applications.
\vspace{-0.3cm}
\subsection{Model Overview}
\vspace{-0.3cm}
Figure~\ref{fig:rewrite-arch} illustrates the overall framework of our neural rewriter, and we describe the two key components for rewriting as follows. More details can be found in
\ifapp
Appendix~\ref{app:model-details}.
\else
\appref.
\fi

\vspace{-0.3cm}
\paragraph{Score predictor.} Given the state $s_t$, the score predictor computes a score $Q(s_t, \omega_t)$ for every $\omega_t \in \Omega(s_t)$, which measures the benefit of rewriting $s_t[\omega_t]$. A high score indicates that rewriting $s_t[\omega_t]$ could be desirable. Note that $\Omega(s_t)$ is a problem-dependent region set. For expression simplification, $\Omega(s_t)$ includes all sub-trees of the expression parse trees; for job scheduling, $\Omega(s_t)$ covers all job nodes for scheduling; and for vehicle routing, it includes all nodes in the route.
\vspace{-0.3cm}
\paragraph{Rule selector.} Given $s_t[\omega_t]$ to be rewritten, the rule-picking policy predicts a probability distribution $\pi_u(s_t[\omega_t])$ over the entire ruleset $\mathcal{U}$, and selects a rule $u_t \in\mathcal{U}$ to apply accordingly.
\vspace{-0.3cm}
\subsection{Training Details}
\vspace{-0.3cm}
Let $(s_0, (\omega_0, u_0)), ..., (s_{T-1}, (\omega_{T-1}, u_{T-1})), s_T$ be the rewriting sequence in the forward pass. 

\textbf{Reward function}. We define $r(s_t, (\omega_t, u_t))$ as $r(s_t, (\omega_t, u_t)) = c(s_t) - c(s_{t+1})$, where $c(\cdot)$ is the task-specific cost function in Section~\ref{sec:setup}.

\textbf{Q-Actor-Critic training.} We train the region-picking policy $\pi_\omega$ and rule-picking policy $\pi_u$ simultaneously. For $\pi_\omega(\omega_t|s_t; \theta)$, we parameterize it as a softmax of the underlying $Q(s_t, \omega_t; \theta)$ function:
\begin{equation}
    \pi_\omega(\omega_t|s_t; \theta) = \frac{\exp(Q(s_t, \omega_t; \theta))}{\sum_{\omega_t} \exp(Q(s_t, \omega_t; \theta))}
\end{equation}
and instead learn $Q(s_t, \omega_t; \theta)$ by fitting it to the cumulative reward sampled from the current policies $\pi_\omega$ and $\pi_u$:
\begin{equation}
    L_\omega(\theta) = \frac{1}{T}\sum_{t=0}^{T-1}(\sum_{t'=t}^{T-1}\gamma^{t'-t}r(s_t', (\omega_t', u_t')) -Q(s_t, \omega_t; \theta))^2
    \label{eq:lsp}
\end{equation}
Where $T$ is the length of the episode (i.e., the number of rewriting steps), and $\gamma$ is the decay factor.

For rule-picking policy $\pi_u(u_t|s_t[\omega_t]; \phi)$, we employ the Advantage Actor-Critic algorithm~\cite{sutton1998reinforcement} with the learned $Q(s_t, \omega_t; \theta)$ as the critic, and thus avoid boot-strapping which could cause sample insufficiency and instability in training. This formulation is similar in spirit to soft-Q learning~\cite{haarnoja2017reinforcement}. Denoting $\Delta(s_t, (\omega_t, u_t)) \equiv \sum_{t'=t}^{T-1}\gamma^{t'-t}r(s_t', (\omega_t', u_t')) - Q(s_t, \omega_t; \theta)$ as the advantage function, the loss function of the rule selector is:
\vspace{-0.15cm}
\begin{equation}
    L_u (\phi) = -\sum_{t=0}^{T-1} \Delta(s_t, (\omega_t, u_t))\log \pi_u(u_t|s_t[\omega_t]; \phi)
    \label{eq:lrs}
\end{equation}
\vspace{-0.2cm}

The overall loss function is $L(\theta, \phi) = L_u(\phi) + \alpha L_\omega(\theta)$, where $\alpha$ is a hyper-parameter. More training details can be found in
\ifapp
Appendix~\ref{app:training-details}.
\else
\appref.
\fi

\begin{figure*}[t]
    \centering
    \includegraphics[width=\textwidth]{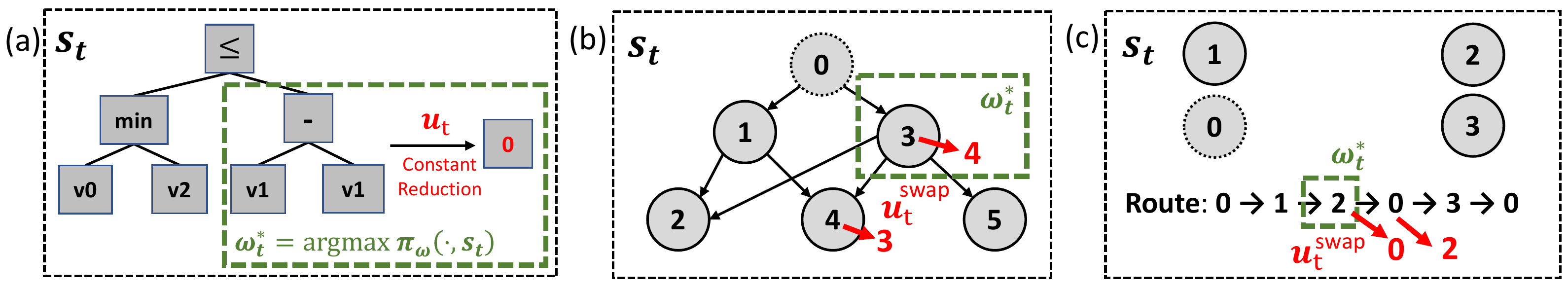}
    \caption{\small{The instantiation of {\ours} for different domains: \textbf{(a)} expression simplification; \textbf{(b)} job scheduling; and \textbf{(c)} vehicle routing. In \textbf{(a)}, $s_t$ is the expression parse tree, where each square represents a node in the tree. The set $\Omega(s_t)$ includes every sub-tree rooted at a non-terminal node, from which the region-picking policy selects $\omega_t\sim \pi_\omega(\omega_t|s_t))$ to rewrite. Afterwards, the rule-picking policy predicts a rewriting rule $u_t \in \mathcal{U}$, then rewrites the sub-tree $\omega_t$ to get the new tree $s_{t+1}$. In \textbf{(b)}, $s_t$ is the dependency graph representation of the job schedule. Each circle with index greater than 0 represents a job node, and node 0 is an additional one representing the machine. Edges in the graph reflect job dependencies. The region-picking policy selects a job $\omega_t$ to re-schedule from all job nodes, then the rule-picking policy chooses a moving action $u_t$ for $\omega_t$, then modifies $s_t$ to get a new dependency graph $s_{t+1}$. In \textbf{(c)}, $s_t$ is the current route, and $\omega_t$ is the node selected to change the visit order. Node 0 is the depot, and other nodes are customers with certain resource demands. The region-picking policy and the rule-picking policy work similarly to the job scheduling ones.}
    }
    \label{fig:rewrite-arch-details}
\end{figure*}

\vspace{-0.3cm}
\section{Applications}
\vspace{-0.3cm}
In the following sections, we discuss the application of our rewriting approach to three different domains: expression simplification, online job scheduling, and vehicle routing. In expression simplification, we minimize the expression length using a well-defined semantics-preserving rewriting ruleset. In online job scheduling, we aim to reduce the overall waiting time of jobs. In vehicle routing, we aim to minimize the total tour length.

\subsection{Expression Simplification}

We first apply our approach to expression simplification domain. In particular, we consider expressions in Halide, a domain-specific language for high-performance image processing~\cite{ragan2013halide}, which is widely used at scale in multiple products of Google (e.g., YouTube) and Adobe Photoshop. Simplifying Halide expressions is an important step towards the optimization of the entire code. To this end, a rule-based rewriter is implemented for the expressions, which is carefully tuned with manually-designed heuristics. The grammar of the expressions considered in the rewriter is specified in
\ifapp
Appendix~\ref{app:Halide-data}.
\else
\appref.
\fi
Notice that the grammar includes a more comprehensive operator set than previous works on finding equivalent expressions, which consider only boolean expressions~\cite{allamanis2017learning,evans2018can} or a subset of algorithmic operations~\cite{allamanis2017learning}. The rewriter includes hundreds of manually-designed rewriting templates. Given an expression, the rewriter checks the templates in a pre-designed order, and applies those rewriting templates that match any sub-expression of the input.

\eat{\yuandong{This paragraph is quite long and detailed. I suggest only delivering main messages here, and putting the details to experiment section. The main message is: rule-based has a lot of pre-conditions, cannot apply up-hill rules properly, and is cumbersome to specify, etc}~\xinyun{Sure, updated!}}

After investigating the rewriting templates in the rule-based rewriter, we find that a large number of rewriting templates enumerate specific cases for an \emph{uphill rule}, which lengthens the expression first and shortens it later (e.g., ``min/max'' expansion). Similar to momentum terms in gradient descent for continuous optimization, such rules are used to escape a local optimum. However, they should only be applied when the initial expression satisfies certain \emph{pre-conditions}, which is traditionally specified by manual design, a cumbersome process that is hard to generalize. 

Observing these limitations, we hypothesize that a neural network model has the potential of doing a better job than the rule-based rewriter. In particular, we propose to only keep the core rewriting rules in the ruleset, remove all unnecessary pre-conditions, and let the neural network decide which and when to apply each rewriting rule. In this way, the neural rewriter has a better flexibility than the rule-based rewriter, because it can learn such rewriting decisions from data, and has the ability of discovering novel rewriting patterns that are not included in the rule-based rewriter.

\textbf{Ruleset}. We incorporate two kinds of templates from Halide rewriting ruleset. The first kind is simple rules (e.g., $v-v \rightarrow 0$), while the second one is the uphill rules after removing their manually designed pre-conditions that do not affect the validity of the rewriting. In this way, a ruleset with $|\mathcal{U}| =19$ categories is built. See
\ifapp
Appendix~\ref{app:ruleset-Halide}
\else
\appref\xspace
\fi
for more details.

\textbf{Model specification.} We use expression parse trees as the input, and employ the N-ary Tree-LSTM designed in~\cite{tai2015improved} as the input encoder to compute the embedding for each node in the tree. Both the score predictor and the rule selector are fully connected neural networks, taken the LSTM embeddings as the input. More details can be found in
\ifapp
Appendix~\ref{app:model-details-Halide}.
\else
\appref.
\fi

\subsection{Job Scheduling Problem}
We also study the job scheduling problem, using the problem setup in~\cite{mao2016resource}.

\yuandonggresolved{For notation, basically I don't want to see $S$, $s$, $r$ since they are reserved by MDP. I prefer using $v$ as each job since they will also be the vertices of the graph.} 

\textbf{Notation}. Suppose we have a machine with $D$ types of resources. Each job $j$ is specified as $v_j = (\boldsymbol{\rho}_j, A_j, T_j)$, where the $D$-dimensional vector $\boldsymbol{\rho}_j = [\rho_{jd}]$ denotes the required portion $0\le\rho_{jd}\le 1$ of the resource type $d$, $A_j$ is the arrival timestep, and $T_j$ is the duration. In addition, we define $B_j$ as the scheduled beginning time, and $C_j = B_j + T_j$ as the completion time. 

We assume that the resource requirement is fixed during the entire job execution, each job must run continuously until finishing, and no preemption is allowed. We adopt an online setting: there is a pending job queue that can hold at most $W$ jobs. When a new job arrives, it can either be allocated immediately, or be added to the queue. If the queue is already full, to make space for the new job, at least one job in the queue needs to be scheduled immediately.
The goal is to find a time schedule for every job, so that the average waiting time is as short as possible. 

\textbf{Ruleset}. The set of rewriting rules is to re-schedule a job $v_j$ and allocate it after another job $v_{j'}$ finishes or at its arrival time $A_j$. See
\ifapp
Appendix~\ref{app:ruleset-jsp}
\else
\appref
\fi
for details of a rewriting step. The size of the rewriting ruleset is $|\mathcal{U}| =2W$, since each job could only switch its scheduling order with at most $W$ of its former and latter jobs respectively.

\textbf{Representation.} We represent each schedule as a directed acyclic graph (DAG), which describes the dependency among the schedule time of different jobs. Specifically, we denote each job $v_j$ as a node in the graph, and we add an additional node $v_0$ to represent the machine. If a job $v_j$ is scheduled at its arrival time $A_j$ (i.e., $B_j = A_j$), then we add a directed edge $\langle v_0, v_j\rangle$ in the graph. Otherwise, there must exist at least one job $v_{j'}$ such that $C_{j'}=B_j$ (i.e., job $j$ starts right after job $j'$). We add an edge $\langle v_{j'}, v_j\rangle$ for every such job $v_{j'}$ to the graph. Figure~\ref{fig:rewrite-arch-details}(b) shows the setting, and we defer the embedding and graph construction details to
\ifapp
Appendix~\ref{app:model-details-jsp}.
\else
\appref.
\fi

\textbf{Model specification}. To encode the graphs, we extend the Child-Sum Tree-LSTM architecture in~\cite{tai2015improved}, which is similar to the DAG-structured LSTM in~\cite{zhu2016dag}. Similar to the expression simplification model, both the score predictor and the rule selector are fully connected neural networks, and we defer the model details to
\ifapp
Appendix~\ref{app:model-details-jsp}.
\else
\appref.
\fi
\subsection{Vehicle Routing Problem}
In addition, we evaluate our approach on vehicle routing problems studied in~\cite{kool2018attention,nazari2018reinforcement}. Specifically, we focus on the Capacitated VRP (CVRP), where a single vehicle with limited capacity needs to satisfy the resource demands of a set of customer nodes. To do so, we construct multiple routes starting and ending at the depot, i.e., node 0 in Figure~\ref{fig:rewrite-arch-details}(c), so that the resources delivered in each route do not exceed the vehicle capacity, while the total route length is minimized.

We represent each vehicle routing problem as a sequence of the nodes visited in the tour, and use a bi-directional LSTM to embed the routes. The ruleset is similar to the job scheduling, where each node can swap with another node in the route. The architectures of the score predictor and rule selector are similar to job scheduling. More details can be found in
\ifapp
Appendix~\ref{app:model-details-vrp}.
\else
\appref.
\fi

\eat{
\subsection{k-SAT}

In addition, we evaluate our approach on k-SAT problems studied in~\cite{amizadeh2018learning,selsam2018learning}. Similar to~\cite{amizadeh2018learning}, we represent each SAT problem as a DAG, where there is a node for each clause and variable, and an edge from every clause to its literals. The model architecture for graph embedding is similar to the job scheduling model, except that we compute the embedding for both the original DAG and the reversed DAG (i.e., the graph with the same set of nodes but reversed edges), and concatenate the node embeddings as the policy input. As shown in Figure~\ref{fig:rewrite-arch-details}, the ruleset includes only one action: flipping the assignment of a variable. Thus, we only need to train the region-picking policy to determine the variable to flip, and do not include the rule-picking policy here.}

%% file: exp.tex
\vspace{-0.3cm}
\section{Experiments} \label{sec:experiments}
\vspace{-0.3cm}
We present the evaluation results in this section. To calculate the inference time, we run all algorithms on the same server equipped with 2 Quadro GP100 GPUs and 80 CPU cores. Only 1 GPU is used when evaluating neural networks, and 4 CPU cores are used for search algorithms. We set the timeout of search algorithms to be 10 seconds per instance. All neural networks in our evaluation are implemented in PyTorch~\cite{paszke2017automatic}.

\def\haliderule{\texttt{Halide-rule}}
\def\zzzrule{\texttt{Z3-simplify}}
\def\zzzsearch{\texttt{Z3-ctx-solver-simplify}}
\def\hsearch{\texttt{Heuristic-search}}

\def\deeprm{\texttt{DeepRM}}
\def\ejf{\texttt{EJF}}
\def\sjf{\texttt{SJF}}
\def\sjfs{\texttt{SJFS}}
\def\sjfo{\texttt{SJF-offline}}

\begin{figure*}
    \centering
    \begin{subfigure}[t]{0.48\linewidth}
        \includegraphics[width=\linewidth]{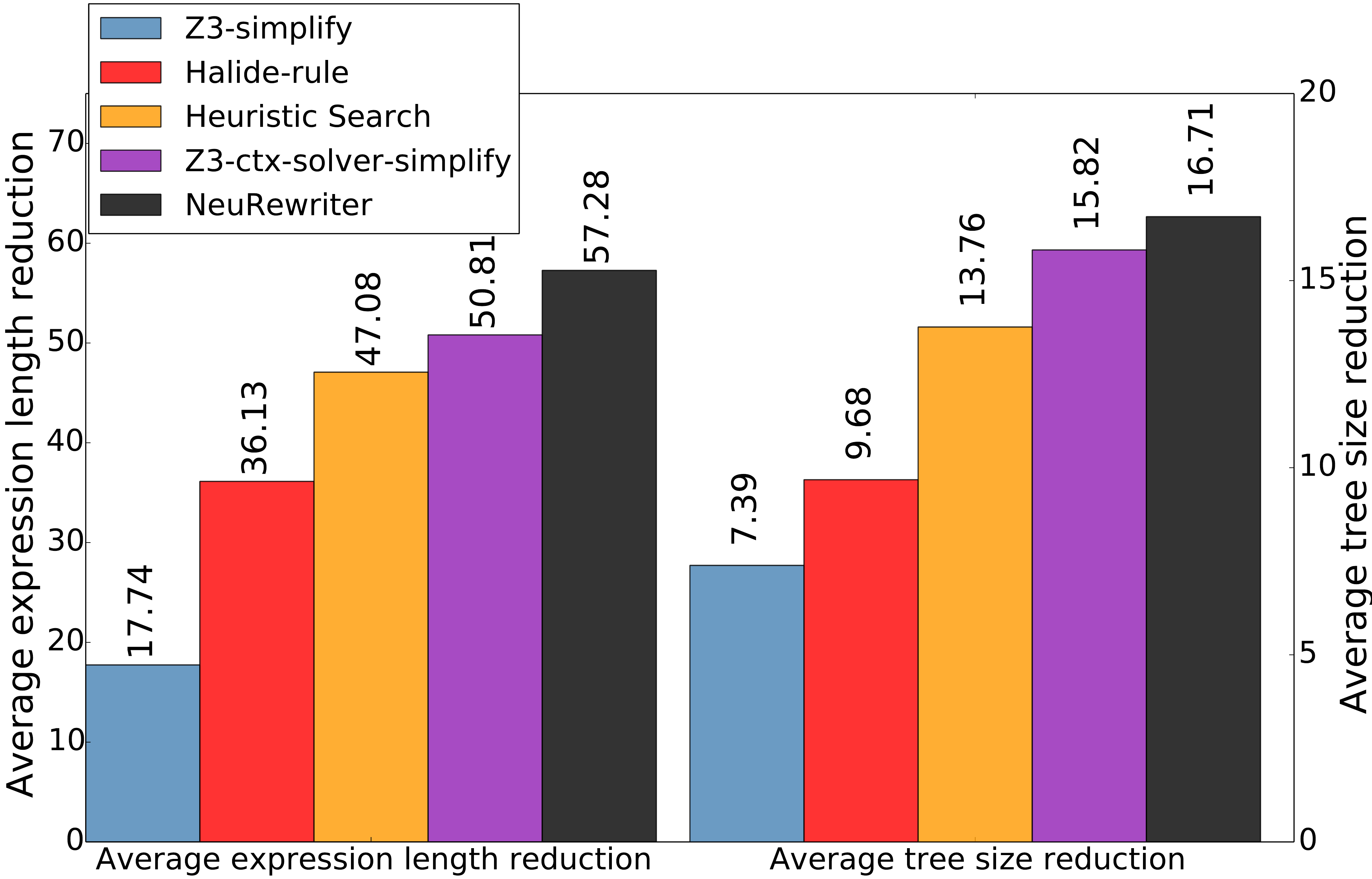}
        \caption{}
        \label{fig:res-Halide-main}
    \end{subfigure}
    \begin{subfigure}[t]{0.48\linewidth}
        \centering
        \includegraphics[width=\linewidth]{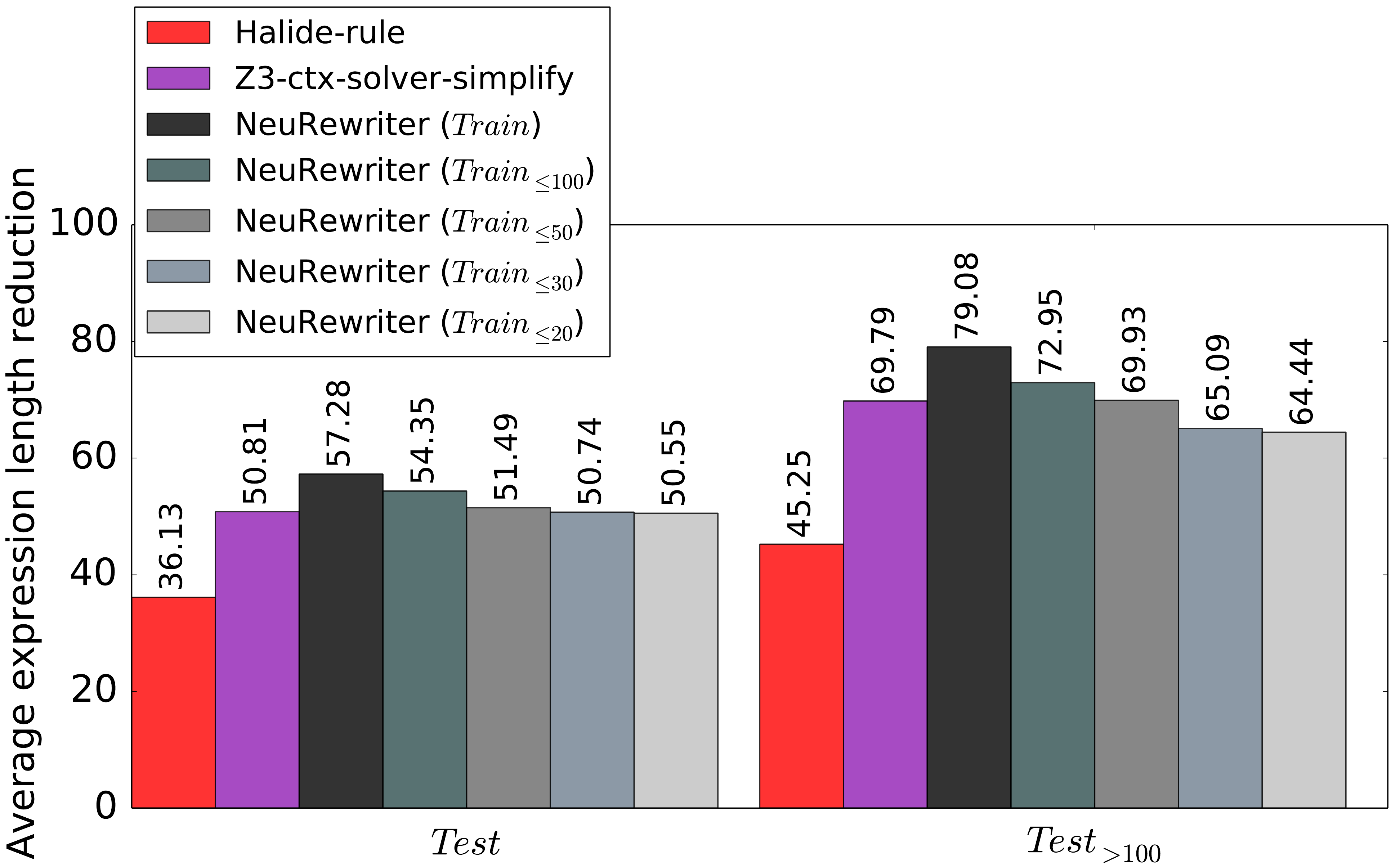}
        \caption{}
        \label{fig:res-Halide-length}
    \end{subfigure}
    \caption{\small{Experimental results of the expression simplification problem. In~\textbf{(b)}, we train {\ours} on expressions of different lengths (described in the brackets).}}
    \label{fig:res-Halide}

    \includegraphics[width=\linewidth]{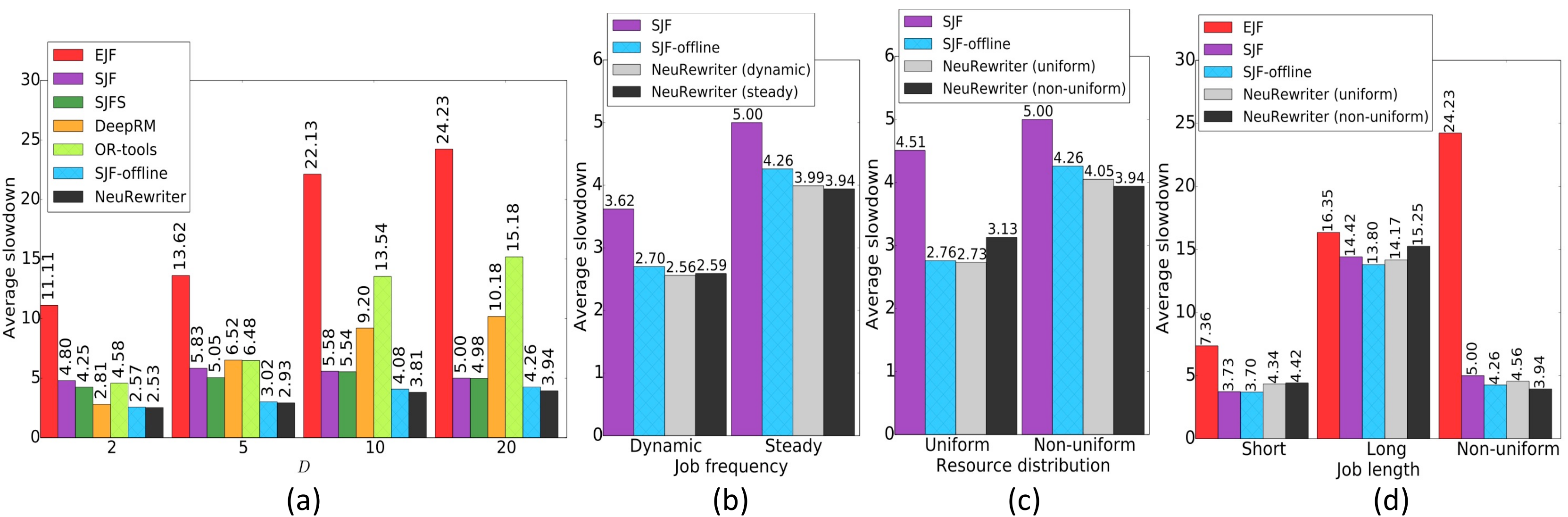}
    \caption{\small{Experimental results of the job scheduling problem varying the following aspects:~\textbf{(a)} the number of resource types $D$; \textbf{(b)} job frequency;~\textbf{(c)} resource distribution;~\textbf{(d)} job length. For {\ours}, we describe training job distributions in the brackets. Workloads in~\textbf{(a)} are with steady job frequency, non-uniform resource distribution, and non-uniform job length. In~\textbf{(b)},~\textbf{(c)} and~\textbf{(d)}, $D=20$. In \textbf{(b)} and~\textbf{(c)}, we omit the comparison with some approaches because their results are significantly worse; for example, the average slowdown of {\ejf} is $14.53$ on the dynamic job frequency, and $11.06$ on the uniform resource distribution. More results can be found in
    \ifapp
    Appendix~\ref{app:jsp-res}.
    \else
    \appref.
    \fi}
    }
    \label{fig:res-jsp}
    
    \begin{subfigure}[t]{0.48\linewidth}
        \includegraphics[width=\linewidth]{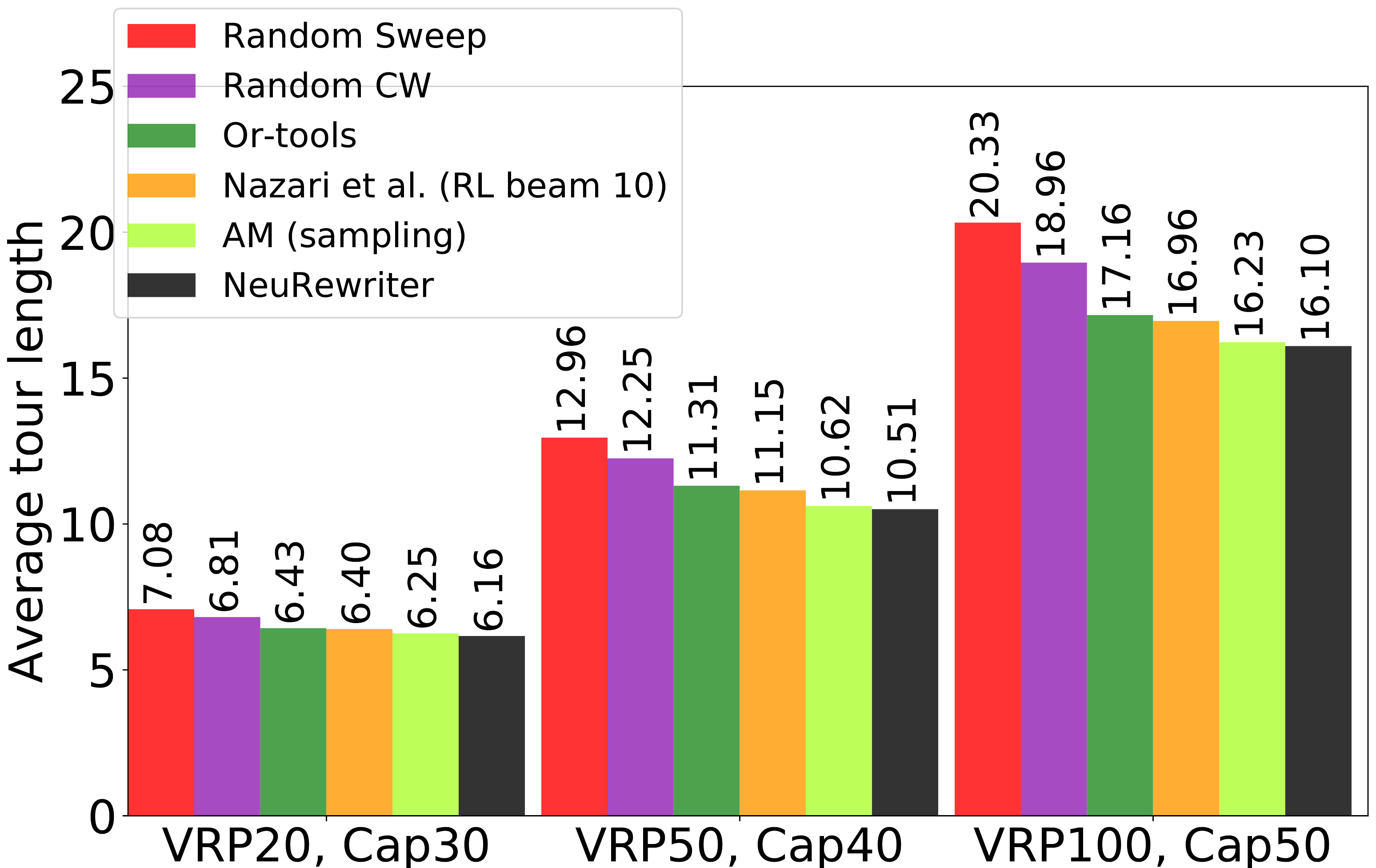}
        \caption{}
        \label{fig:res-vrp-main}
    \end{subfigure}
    \begin{subfigure}[t]{0.48\linewidth}
        \centering
        \includegraphics[width=\linewidth]{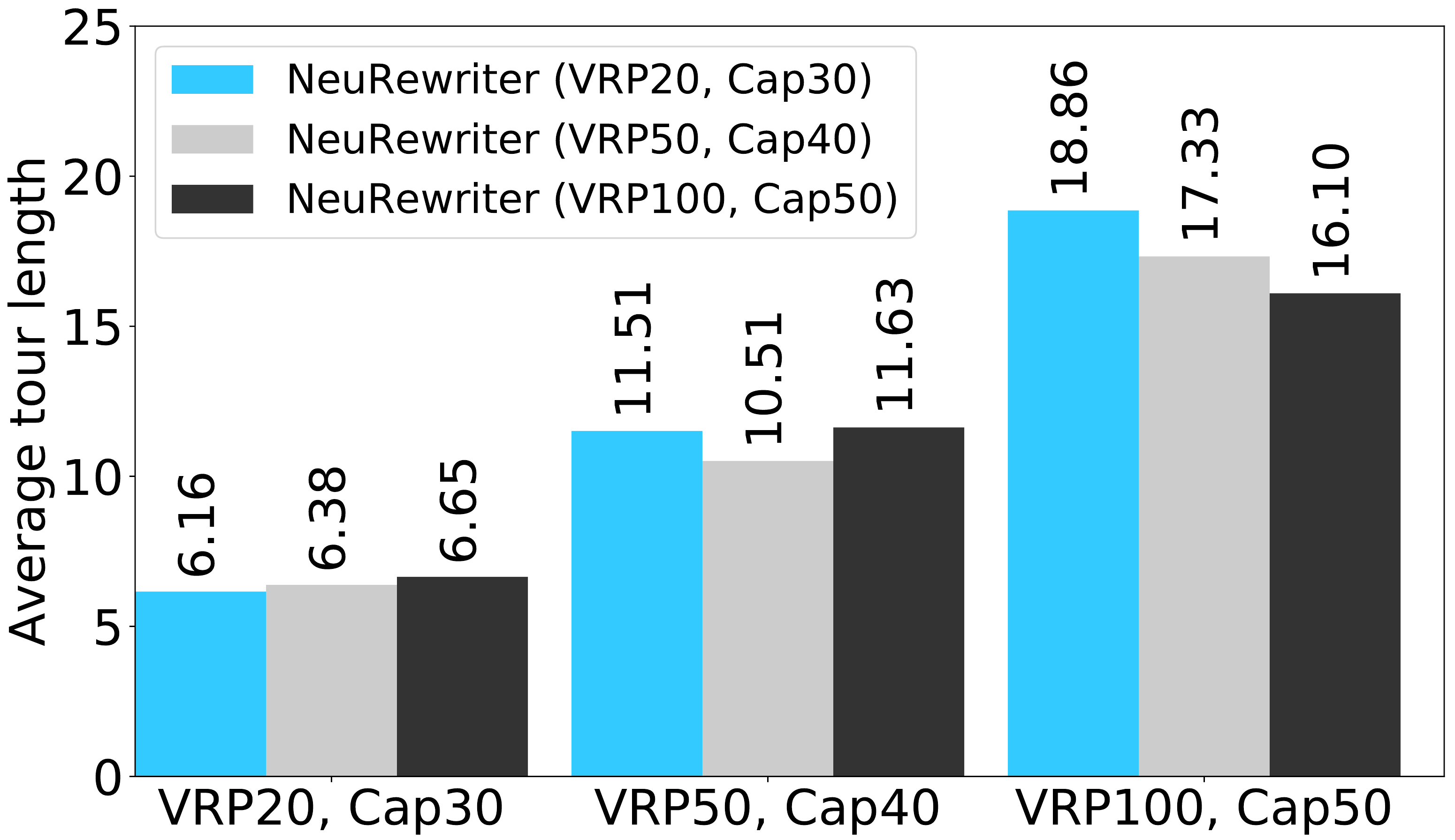}
        \caption{}
        \label{fig:res-vrp-gen}
    \end{subfigure}
    \caption{\small{Experimental results of the vehicle routing problem with different number of customer nodes and vehicle capacity; e.g., VRP100, Cap50 means there are 100 customer nodes and the vehicle capacity is 50. \textbf{(a)} {\ours} outperforms multiple baselines and previous works~\cite{kool2018attention, nazari2018reinforcement}. More results can be found in
    \ifapp
    Appendix~\ref{app:vrp-res}.
    \else
    \appref.
    \fi
    \textbf{(b)} We evaluate the generalization performance of {\ours} on problems from different distributions, and we describe the training problem distributions in the brackets.}}
    \label{fig:res-vrp}
\end{figure*}
\vspace{-0.3cm}
\subsection{Expression Simplification}
\textbf{Setup}. To construct the dataset, we first generate random pipelines using the generator in Halide, then extract expressions from them.
We filter out those irreducible expressions, then split the rest into 8/1/1 for training/validation/test sets respectively. See
\ifapp
Appendix~\ref{app:Halide-data}
\else
\appref\xspace
\fi 
for more details.

\textbf{Metrics}. We evaluate the following two metrics: (1)~\emph{Average expression length reduction}, which is the length reduced from the initial expression to the rewritten one, and the length is defined as the number of characters in the expression; (2)~\emph{Average tree size reduction}, which is the number of nodes decreased from the initial expression parse tree to the rewritten one.

\eat{
\begin{table*}[t]
    \centering
    \begin{tabular}{|c|c|c|}
    \hline
    & Average expression length reduction & Average tree size reduction \\
    \hline
    \zzzrule{} & 17.74 & 7.39 \\
    \haliderule{} & 36.13 & 9.68 \\
    Heuristic Search & 47.08 & 13.76 \\
    \zzzsearch{} & 50.81 & 15.82 \\
    \hline\hline
    \ours & \textbf{57.28} & \textbf{16.71} \\
    \hline
    \end{tabular}
    \caption{Experimental results of the Halide expression simplification task on the test set. The Z3 simplifier (rule-based) refers to the ``simplify'' tactic in Z3, and the Z3 simplifier (solver-based) refers to the ``ctx-solver-simplify'' tactic. Note that the Z3 solver-based simplifier can perform rewriting steps that are not included in the Halide ruleset.\yuandong{Also convert the tables to figures. Similar as above.}}
    \label{tab:res-Halide}
\end{table*}}

\textbf{Baselines}. We examine the effectiveness of {\ours} against two kinds of baselines. The first kind of baselines are heuristic-based rewriting approaches, including {\haliderule} (the rule-based Halide rewriter in Section~\ref{sec:setup}) and \hsearch, which applies beam search to find the shortest rewriting with our ruleset at each step. Note that \ours{} does not use beam search. 

In addition, we also compare our approach with Z3, a high-performance theorem prover developed by Microsoft Research~\cite{de2008z3}. Z3 provides two tactics to simplify the expressions: \zzzrule{} performs some local transformation using its pre-defined rules, and \zzzsearch{} traverses each sub-formula in the input expression and invokes the solver to find a simpler equivalent one to replace it. This search-based tactic is able to perform simplification not included in the Halide ruleset, and is generally better than the rule-based counterpart but with more computation. For \zzzsearch{}, we set the timeout to be 10 seconds for each input expression.

\textbf{Results}. Figure~\ref{fig:res-Halide-main} presents the main results. We can notice that the performance of \zzzrule{} is worse than \haliderule{}, because the ruleset included in this simplifier is more restricted than the Halide one, and in particular, it can not handle expressions with ``max/min/select'' operators. On the other hand, {\ours} outperforms both the rule-based rewriters and the heuristic search by a large margin. In particular, {\ours} could reduce the expression length and parse tree size by around $52\%$ and $59\%$ on average; compared to the rule-based rewriters, our model further reduces the average expression length and tree size by around $20\%$ and $15\%$ respectively. We observe that the main performance gain comes from learning to apply uphill rules appropriately in ways that are not included in the manually-designed templates. For example, consider the expression $5 \leq \max(\max(v0, 3) + 3, \max(v1, v2))$, which could be reduced to $True$ by expanding $\max(\max(v0, 3) + 3, \max(v1, v2))$ and $\max(v0, 3)$. Using a rule-based rewriter would require the need of specifying the pre-conditions recursively, which becomes prohibitive when the expressions become more complex. On the other hand, heuristic search may not be able to find the correct order of expanding the right hand size of the expression when more ``min/max'' are included, which would make the search less efficient.

Furthermore, {\ours} also outperforms \zzzsearch{} in terms of both the result quality and the time efficiency, as shown in Figure~\ref{fig:res-Halide-main} and Table~\ref{tab:time-expr}. Note that the implementation of Z3 is in C++ and highly optimized, while {\ours} is implemented in Python; meanwhile, \zzzsearch{} could perform rewriting steps that are not included in the Halide ruleset. More results can be found in
\ifapp
Appendix~\ref{app:Halide-res}.
\else
\appref.
\fi

\def\train#1{\emph{Train}$_{\le #1}$}
\def\test#1{\emph{Test}$_{> #1}$}

\textbf{Generalization to longer expressions}. To measure the generalizability of our approach, we construct 4 subsets of the training set: \train{20}, \train{30}, \train{50} and \train{100}, which only include expressions of length at most 20, 30, 50 and 100 in the full training set. We also build~\test{100}, a subset of the full test set that only includes expressions of length larger than 100. The statistics of these datasets can be found in
\ifapp
Appendix~\ref{app:Halide-data}.
\else
\appref.
\fi

We present the results of training our model on different datasets above in Figure~\ref{fig:res-Halide-length}. Even trained on short expressions, \ours{} is still comparable with the Z3 solver. Thanks to local rewriting rules, our approach can generalize well even when operating on very different data distributions. \yuandong{We might need to put one concrete example on how the expression is simplified (and how Z3 failed).}

\eat{
\begin{table*}[t]
    \centering
    \begin{tabular}{|c|c|c|c|}
    \hline
    \multicolumn{2}{|c|}{} & Test & \test{100} \\
    \hline
    \multicolumn{2}{|c|}{\haliderule{}} & 36.13 & 45.25 \\
    \multicolumn{2}{|c|}{\zzzsearch{}} & 50.81 & 69.79 \\
    \hline
    \multirow{5}{*}{\ours} & Train & \textbf{57.28} & \textbf{79.08} \\
    & \train{100} & 54.35 & 72.95\\
    & \train{50} & 51.49 & 69.93 \\
    & \train{30} & 50.74 & 65.09 \\
    & \train{20} & 50.55 & 64.44 \\
    \hline
    \end{tabular}
    \caption{Experimental results of the neural rewriter trained on expressions of different lengths. Here,~\emph{Test} is the full testset,~\test{100} is the subset of expressions of length greater than 100,~\emph{Train} is the full training set,~\train{100},~\train{50},~\train{30} and~\train{20} are subsets including expressions of length at most 100, 50, 30 and 20 respectively. We report the average expression length reduction for both testsets.}
    \label{tab:res-Halide-length}
\end{table*}}
\vspace{-0.3cm}
\subsection{Job Scheduling Problem}
\textbf{Setup}. We randomly generate $100K$ job sequences, and use $80K/10K/10K$ for training, validation and testing. Typically each job sequence includes $\sim 50$ jobs. We use an online setting where jobs arrive on the fly with a pending job queue of length $W=10$. Unless stated otherwise, we generate initial schedules using \emph{Earliest Job First} (\ejf), which can be constructed with negligible overhead. 

When the number of resource types $D=2$, we follow the same setup as in~\cite{mao2016resource}. The maximal job duration $T_{max}=15$, and the latest job arrival time $A_{max}=50$. With larger $D$, except changing the resource requirement of each job to include more resource types, other configurations stay the same. 

\textbf{Metric}. Following \deeprm ~\cite{mao2016resource}, we use the~\emph{average job slowdown} $\eta_j \equiv (C_j - A_j) / T_j \ge 1$ as our evaluation metric. Note that $\eta_j = 1$ means no slow down.

\textbf{Job properties.} To test the stability and generalizability of \ours, we change job properties (and their distributions): (1) \emph{Number of resource types $D$}: larger $D$ leads to more complicated scheduling; (2) \emph{Average job arrival rate}: the probability that a new job will arrive, \emph{Steady job frequency} sets it to be $70\%$, and \emph{Dynamic job frequency} means the job arrival rate changes randomly at each timestep; (3) \emph{Resource distribution}: jobs might require different resources, where some are \emph{uniform} (e.g., half-half for resource 1 and 2) while others are \emph{non-uniform} (see
    \ifapp
    Appendix~\ref{app:jsp-data}
    \else
    \appref\xspace
    \fi 
    for the detailed description);
    (4) \emph{Job lengths}: \emph{Uniform job length}: length of each job in the workload is either $[10, 15]$ (long) or $[1, 3]$ (short), and \emph{Non-uniform job length}: workload has both short and long jobs. We show that {\ours} is fairly robust under different distributions. When trained on one distribution, it can generalize to others without performance collapse. 

We compare {\ours} with three kinds of baselines. 

\textbf{Baselines on Manually designed heuristics}: \textit{Earliest Job First} (\ejf) schedules each job in the increasing order of their arrival time. \textit{Shortest Job First} (\sjf) always allocates the shortest job in the pending job queue at each timestep, which is also used as a baseline in~\cite{mao2016resource}. \textit{Shortest First Search} (\sjfs) searches over the shortest $k$ jobs to schedule at each timestep, and returns the optimal one. We find that other heuristic-based baselines used in~\cite{mao2016resource} generally perform worse than \sjf, especially with large $D$. Thus, we omit the comparison. 

\textbf{Baselines on Neural network.} We compare with \deeprm~\cite{mao2016resource}, a neural network also trained with RL to construct a solution from scratch.

\textbf{Baselines on Offline planning}. To measure the optimality of these algorithms, we also take an offline setting, where the entire job sequence is available before scheduling. Note that this is equivalent to assuming an unbounded length of the pending job queue. With such additional knowledge, this setting provides a strong baseline. We tried two offline algorithms: (1) {\sjfo}, which is a simple heuristic that schedules each job in the increasing order of its duration; and (2) Google OR-tools~\cite{GoogleOrTools}, which is a generic toolbox for combinatorial optimization. For OR-tools, we set the timeout to be 10 seconds per workload, but we find that it can not achieve a good performance even with a larger timeout, and we defer the discussion to
\ifapp
Appendix~\ref{app:jsp-res}.
\else
\appref.
\fi
\label{sec:eval-job-scheduling-main}

\textbf{Results on Scalability}. As shown in Figure~\ref{fig:res-jsp}a, {\ours} outperforms both heuristic algorithms and the baseline neural network \deeprm. In particular, while the performance of {\deeprm} and {\ours} are similar when $D=2$, with larger $D$, {\deeprm} starts to perform worse than heuristic-based algorithms, which is consistent with our hypothesis that it becomes challenging to design a schedule from scratch when the environment becomes more complex. On the other hand, {\ours} could capture the bottleneck of an existing schedule that limits its efficiency, then progressively refine it to obtain a better one. In particular, our results are even better than offline algorithms that assume the knowledge of the entire job sequence, which further demonstrates the effectiveness of {\ours}. Meanwhile, we present the running time of OR-tools, {\deeprm} and {\ours} in Table~\ref{tab:time-jsp}. We can observe that both {\deeprm} and {\ours} are much more time-efficient than OR-tools; on the other hand, the running time of {\ours} is comparable to {\deeprm}, while achieving much better results. More discussion can be found in
\ifapp
Appendix~\ref{app:jsp-res}.
\else
\appref.
\fi
\yuandong{We need to add Google OR results as well, which is also a good point for our scalability arguments. Looks like the constraint of ``no overlap among interval variables'' makes Google OR-tools a bit slow, and also even OR-tools gives result, it is not better.}

\textbf{Results on Robustness}. As shown in Figure~\ref{fig:res-jsp}, {\ours} excels in almost all different job distributions, except when the job lengths are uniform (short or long, Figure~\ref{fig:res-jsp}d), in which case existing methods/heuristics are sufficient. This shows that {\ours} can deal with complicated scenarios and is adaptive to different distributions. 

\textbf{Results on Generalization.} Furthermore, {\ours} can also generalize to different distributions than those used in training, without substantial performance drop. This shows the power of local rewriting rules: using local context could yield more generalizable solutions\yuandong{Need to rewrite this sentence (or more experiments to justify that. E.g, the context of neural network is larger/smaller?)}. 

\subsection{Vehicle Routing Problem}
\textbf{Setup and Baselines.} We follow the same training setup as~\cite{kool2018attention,nazari2018reinforcement} by randomly generating vehicle routing problems with different number of customer nodes and vehicle capacity. We compare with two neural network approaches, i.e., AM~\cite{kool2018attention} and Nazari et al.~\cite{nazari2018reinforcement}, and both of them train a neural network policy using reinforcement learning to construct the route from scratch. We also compare with OR-tools and several classic heuristics studied in~\cite{nazari2018reinforcement}. 

\textbf{Results.} We first demonstrate our main results in Figure~\ref{fig:res-vrp-main}, where we include the variant of each baseline that performs the best, and defer more results to \ifapp
Appendix~\ref{app:vrp-res}.
\else
\appref.
\fi Note that the initial routes generated for {\ours} are even worse than the classic heuristics; however, starting from such sub-optimal solutions, {\ours} is still able to iteratively improve the solutions and outperforms all the baseline approaches on different problem distributions. In addition, for VRP20 problems, we can compute the exact optimal solutions, which provides an average tour length of 6.10. We observe that the result of {\ours} (i.e., 6.16) is the closest to this lower bound, which also demonstrates that {\ours} is able to find solutions with better quality.

We also compare the runtime of the most competitive approaches in Table~\ref{tab:time-vrp}. Note that the OR-Tools solver for vehicle routing problems is highly tuned and implemented in C++, while the RL-based approaches in comparison are implemented in Python. Meanwhile, following~\cite{nazari2018reinforcement}, to report the runtime of RL models, we decode a single instance at a time, thus there is potential room for speed improvement by decoding multiple instances per batch. Nevertheless, we can still observe that {\ours} achieves a better balance between the result quality and the time efficiency, especially with a larger problem scale.

\textbf{Results on Generalization.} Furthermore, in Figure~\ref{fig:res-vrp-gen}, we show that {\ours} can generalize to different problem distributions than training ones. In particular, they still exceed the performance of the classic heuristics, and are sometimes comparable or even better than the OR-tools. More discussion can be found in
\ifapp
Appendix~\ref{app:vrp-res}.
\else
\appref.
\fi

\begin{table}[t]
    \begin{subtable}[t]{0.25\textwidth}
    \centering
    \scalebox{0.85}{\begin{tabular}{|c|c|}
    \hline
     & Time (s) \\
    \hline
    Z3-solver & 1.375 \\
    NeuRewriter & 0.159 \\
    \hline
    \end{tabular}}
    \caption{}
    \label{tab:time-expr}
    \end{subtable}
    \begin{subtable}[t]{0.28\textwidth}
    \centering
    \scalebox{0.85}{\begin{tabular}{|c|c|}
    \hline
     & Time (s) \\
    \hline
    OR-tools & 10.0 \\
    DeepRM & 0.020 \\
    NeuRewriter & 0.037 \\
    \hline
    \end{tabular}}
    \caption{}
    \label{tab:time-jsp}
    \end{subtable}
    \begin{subtable}[t]{0.3\textwidth}
    \centering
    \scalebox{0.85}{\begin{tabular}{|c|c|c|c|}
    \hline
    & VRP20 & VRP50 & VRP100 \\
    \hline
    OR-tools & 0.010 & 0.053 & 0.231 \\
    Nazari et al. & 0.162 & 0.232 & 0.445 \\
    AM & 0.036 & 0.168 & 0.720 \\
    NeuRewriter & 0.133 & 0.211 & 0.398 \\
    \hline
    \end{tabular}}
    \caption{}
    \label{tab:time-vrp}
    \end{subtable}
    \vspace{-0.5em}
    \caption{\small{Average runtime (per instance) of different solvers (OR-tools~\cite{GoogleOrTools} and the tactic \zzzsearch{} of Z3~\cite{de2008z3}) and RL-based approaches ({\ours}, DeepRM~\cite{mao2016resource}, Nazari et al.~\cite{nazari2018reinforcement} and AM~\cite{kool2018attention}) over the test set of: \textbf{(a)} expression simplification; \textbf{(b)} job scheduling; \textbf{(c)} vehicle routing.}}
    \vspace{-1em}
    \label{tab:runtime}
\end{table}

%% file: conc.tex
\vspace{-0.3cm}
\section{Conclusion}
\vspace{-0.3cm}
\label{sec:conc}

In this work, we propose to formulate optimization as a rewriting problem, and solve the problem by iteratively rewriting an existing solution towards the optimum. We utilize deep reinforcement learning to train our neural rewriter. In our evaluation, we demonstrate the effectiveness of our neural rewriter on multiple domains, where our model outperforms both heuristic-based algorithms and baseline deep neural networks that generate an entire solution directly.

Meanwhile, we observe that since our approach is based on local rewriting, it could become time-consuming when large changes are needed in each iteration of rewriting. In extreme cases where each rewriting step needs to change the global structure, starting from scratch becomes preferrable. We consider improving the efficiency of our rewriting approach and extending it to more complicated scenarios as future work.

%% file: app.tex
\section{More Details of the Dataset}
\label{app:data-details}

\subsection{Expression Simplification}
\label{app:Halide-data}
\yuandong{Note that we need to put appendix to supplementary materials}
Figure~\ref{fig:Halide-grammar} presents the grammar of Halide expressions in our evaluation. We use the random pipeline generator in the Halide repository to build the dataset~\footnote{\url{https://github.com/halide/Halide/tree/new_autoschedule_with_new_simplifier/apps/random_pipeline}.}. Table~\ref{tab:Halide-stat} presents the statistics of the datasets.

\begin{figure*}[t]
$$
\begin{array}{rcl}
    \texttt{<Expr>} & ::= & \texttt{<AlgExpr>} \mid \texttt{<BoolExpr>} \\
    \texttt{<BoolExpr>} & ::= & \texttt{<AlgExpr>} \textrm{ \textless } \texttt{<AlgExpr>} \\
    & | & \texttt{<AlgExpr>} \textrm{ \textless= } \texttt{<AlgExpr>} \\
    & | & \texttt{<AlgExpr>} \textrm{ == } \texttt{<AlgExpr>} \\
    & | & (!\texttt{<BoolExpr>})\\
    & | & (\texttt{<BoolExpr>} \textrm{ \&\& } \texttt{<BoolExpr>})\\
    & | & (\texttt{<BoolExpr>} \textrm{ || } \texttt{<BoolExpr>})\\
    \texttt{<AlgExpr>} & ::= & \texttt{<Term>} \\
    & | & (\texttt{<AlgExpr>} \textrm{ + } \texttt{<Term>}) \\
    & | & (\texttt{<AlgExpr>} \textrm{ - } \texttt{<Term>}) \\
    & | & (\texttt{<AlgExpr>} \textrm{ * } \texttt{<Term>}) \\
    & | & (\texttt{<AlgExpr>} \textrm{ / } \texttt{<Term>}) \\
    & | & (\texttt{<AlgExpr>} \textrm{ \% } \texttt{<Term>}) \\
    \texttt{<Term>} & ::= & \texttt{<Var>} \mid \texttt{<Const>} \\
    & | & \textbf{max} (\texttt{<AlgExpr>}, \texttt{<AlgExpr>}) \\
    & | & \textbf{min} (\texttt{<AlgExpr>}, \texttt{<AlgExpr>}) \\
    & | & \textbf{select} (\texttt{<BoolExpr>}, \texttt{<AlgExpr>}, \texttt{<AlgExpr>}) \\
    
\end{array}
$$
\caption{Grammar of the Halide expressions in our evaluation. ``select ($c$, $e1$, $e2$)'' means that when the condition $c$ is satisfied, this term is equal to $e1$, otherwise is equal to $e2$. In our dataset, all constants are integers ranging in $[-1024, 1024]$, and variables are from the set $\{v0, v1, ..., v12\}$.}
\label{fig:Halide-grammar}
\end{figure*}

\ifappsubmit
\def\train#1{\emph{Train}$_{\le #1}$}
\def\test#1{\emph{Test}$_{> #1}$}
\fi

\begin{table*}[t]
    \centering
    \begin{tabular}{|c|c|c|}
        \hline
         Number of expressions in the dataset & Length of expressions & Size of expression parse trees \\
         \hline
         Total: 1.36M & Average: 106.84 & Average: 27.39 \\
         Train/Val/Test: 1.09M/136K/136K & Min/Max: 10/579 & Min/Max:3/100 \\
         \train{20}: 17K & Average: 16.76 & Average: 4.66 \\
         \train{30}: 48K & Average: 22.91 & Average: 6.43 \\
         \train{50}: 170K & Average: 35.62 & Average: 10.18 \\
         \train{100}: 588K & Average: 63.49 & Average: 18.72 \\
         \test{100}: 53K & Average: 142.22 & Average: 42.20 \\
         \hline
    \end{tabular}
    \caption{Statistics of the dataset for expression simplification.}
    \label{tab:Halide-stat}
\end{table*}

\subsection{Job Scheduling}
\label{app:jsp-data}

\paragraph{Description of different resource distributions.} For each job $j$, we define~\emph{dominant resources} $d_\text{dom}$ as the resources with $0.5 \leq \rho_{jd_\text{dom}} \leq 1$, and~\emph{auxiliary resources} $d_\text{aux}$ as those with $0.1 \leq \rho_{jd_\text{aux}} \leq 0.2$. We refer to a job with both dominant and auxiliary resources as a job with~\emph{non-uniform resources}. We also evaluate on workloads including only jobs with~\emph{uniform resources}, where each job only includes either dominant resources or auxiliary resources.

\subsection{Vehicle Routing}

Our data generation follows the setup in~\cite{nazari2018reinforcement,kool2018attention}. The positions of the depot and customer nodes are uniformly randomly sampled from the unit square $[0, 1]\times [0, 1]$. Each node is denoted as $v_j = ((x_j, y_j), \delta_j)$, where $(x_j, y_j)$ is the position, and $\delta_j$ is the resource demand. We set $\delta_0=0$ for the depot (i.e., node 0), and $\delta_j \in \{1, 2, ..., 9\}$ for customer nodes (i.e., $j > 0$).

\section{More Details on the Rewriting Ruleset}
\label{app:ruleset}

\subsection{More Details for Expression Simplification Problem}
\label{app:ruleset-Halide}

The ruleset implemented in the Halide rule-based rewriter can be found in their public repository~\footnote{~\url{https://github.com/halide/Halide}.}.

\begin{figure*}[t]
    \centering
    \includegraphics[width=0.7\linewidth]{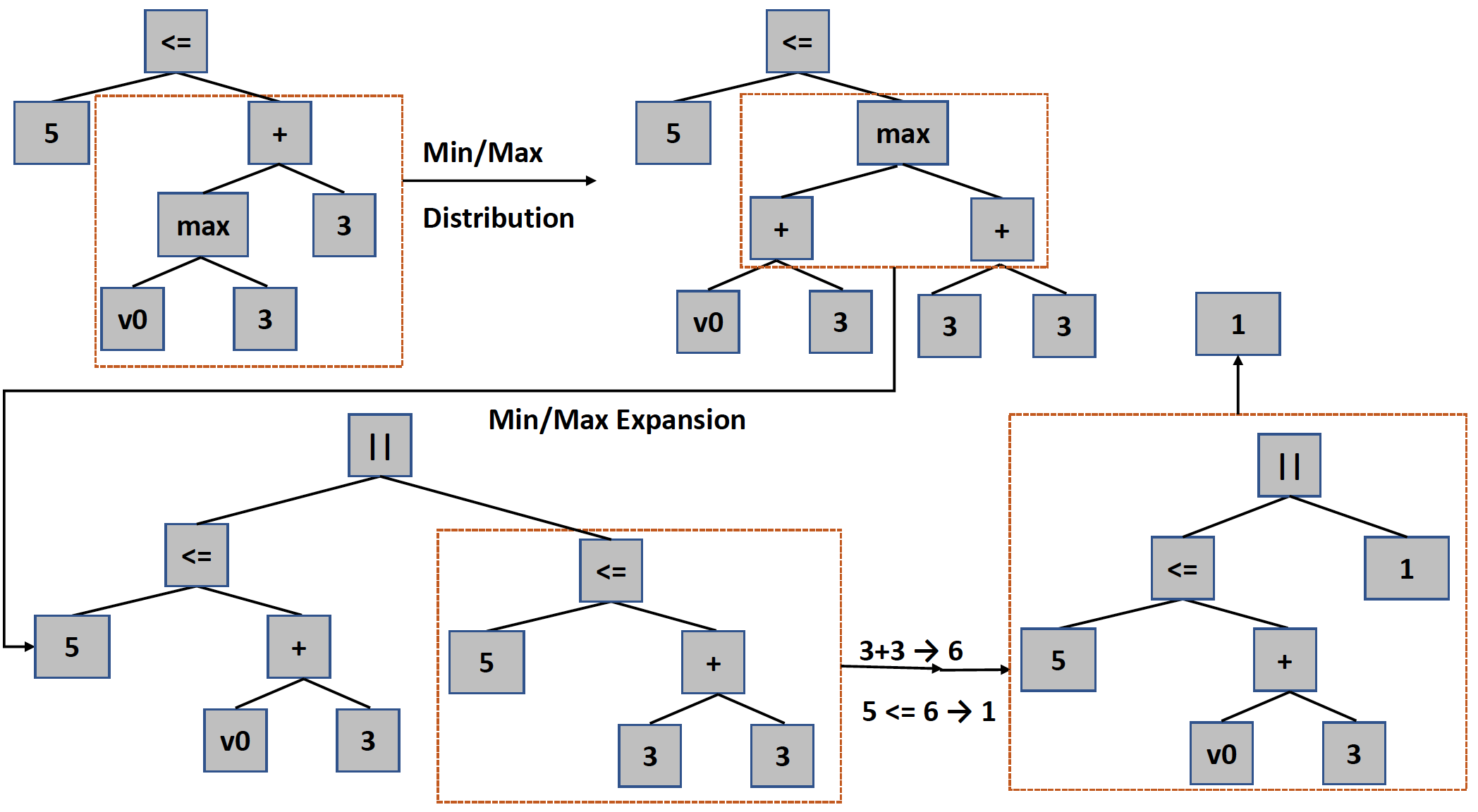}
    \caption{An example of the rewriting process for Halide expressions. The initial expression is $5 \leq max(v0, 3) + 3$, which could be reduced to $1$, i.e., $True$.}
    \label{fig:Halide-rewrite-process}
\end{figure*}

\paragraph{More discussions about the uphill rules.} A commonly used type of uphill rules is ``min/max'' expansion, e.g., $\texttt{min}(a,b) < c \to a < c || b < c$. Dozens of templates in the ruleset of the Halide rewriter are describing conditions when a ``min/max'' expression could be simplified. Notice that although applying this rewriting rule has no benefit in most cases, since it will increase the expression length, it is necessary to include it in the ruleset, because when either $a < c$ or $b < c$ is always true, expanding the ``min'' term could reduce the entire expression to a tautology, which ends up simplifying the entire expression. Figure~\ref{fig:Halide-rewrite-process} shows an example of the rewriting process using uphill rules properly.

\subsection{More Details for Job Scheduling Problem}
\label{app:ruleset-jsp}

Algorithm~\ref{alg:rewrite-jsp} describes a single rewriting step for job scheduling problem.

\begin{algorithm*}[t]
\caption{Algorithm of a Single Rewriting Step for Job Scheduling Problem}
\label{alg:rewrite-jsp}
\begin{algorithmic}[1]
\Function{Rewrite}{$v_j, v_{j'}, s_t$}
\If {$C_{j'} < A_j$ or $C_{j'} == B_j$}
    \State \Return $s_t$
\EndIf
\State $\ifes{j' \neq 0}{B'_j = C_{j'}}{B'_j = A_j}$
\State $C'_j = B'_j + T_j$
\State
\State //Resolve potential resource occupation overflow within $[B'_j, C'_j]$
\State $J = $ all jobs in $s_t$ except $v_j$ that are scheduled within $[B'_j, C'_j]$
\State Sort $J$ in the topological order
\For {$v_i \in J$}
    \State $B'_i = $ the earliest time that job $v_i$ can be scheduled
    \State $C'_i = B'_i + T_i$
\EndFor
\State For $v_i\not\in J$, $B'_i = B_i$, $C'_i = C_i$
\State $s_{t+1} = \{(B'_i, C'_i)\}$
\State \Return $s_{t+1}$
\EndFunction
\end{algorithmic}
\end{algorithm*}

\section{More Details on Model Architectures}
\label{app:model-details}

\subsection{Model Details for Expression Simplification}
\label{app:model-details-Halide}

\paragraph{Input embedding.} Notice that in this problem, each non-terminal has at most 3 children. Thus, let $x$ be the embedding of a non-terminal, $(h_L, c_L), (h_M, c_M), (h_R, c_R)$ be the LSTM states maintained by its children nodes, the LSTM state of the non-terminal node is computed as
\begin{equation}
    (h, c) = \mathrm{LSTM} (([h_L; h_M; h_R], [c_L; c_M; c_R]), x)
    \label{eq:Halide-LSTM}
\end{equation}
Where $[a;b]$ denotes the concatenation of vectors $a$ and $b$. For non-terminals with less than 3 children, the corresponding LSTM states are set to be zero. We use $d$ to represent the size of $h$ and $c$, i.e., the hidden size of the LSTM.

\paragraph{Input representation.} For each sub-tree $\omega_i$, its input to both the score predictor and the rule-picking policy is represented as a $2d$-dimensional vector $[h_0; h_i]$, where $h_0$ is the embedding of the root node encoding the entire tree. The reason why we include $h_0$ in the input is that looking at the sub-tree itself is sometimes insufficient to determine whether it is beneficial to perform the rewriting. For example, consider the expression
$max(a, b) + 2 < a + 2$, by looking at the sub-expression $max(a, b) + 2$ itself, it does not seem necessary to rewrite it as $max (a + 2, b + 2)$. However, given the entire expression, we can observe that this rewriting is an important step towards the simplification, since the resulted expression
$max(a + 2, b + 2) < a + 2$ could be reduced to $False$. We have tried other approaches of combining the parent information into the input, but we find that including the embedding of the entire tree is the most efficient way.

\paragraph{Score predictor.} The score predictor is an $L_P$-layer fully connected network with a hidden size of $N_P$. For each sub-tree $\omega_i$, its input to the score predictor is represented as a $2d$-dimensional vector $[h_0; h_i]$, where $h_0$ embeds the entire tree.

\paragraph{Rule selector.} The rule selector is an $L_S$-layer fully connected network with the hidden size $N_S$, and its input format is the same as the score predictor. A $|\mathcal{U}|$-dimensional softmax layer is used as the output layer.

\subsection{More Details for Job Scheduling Problem}
\label{app:model-details-jsp}

\begin{figure*}[t]
    \centering
    \includegraphics[width=0.7\linewidth]{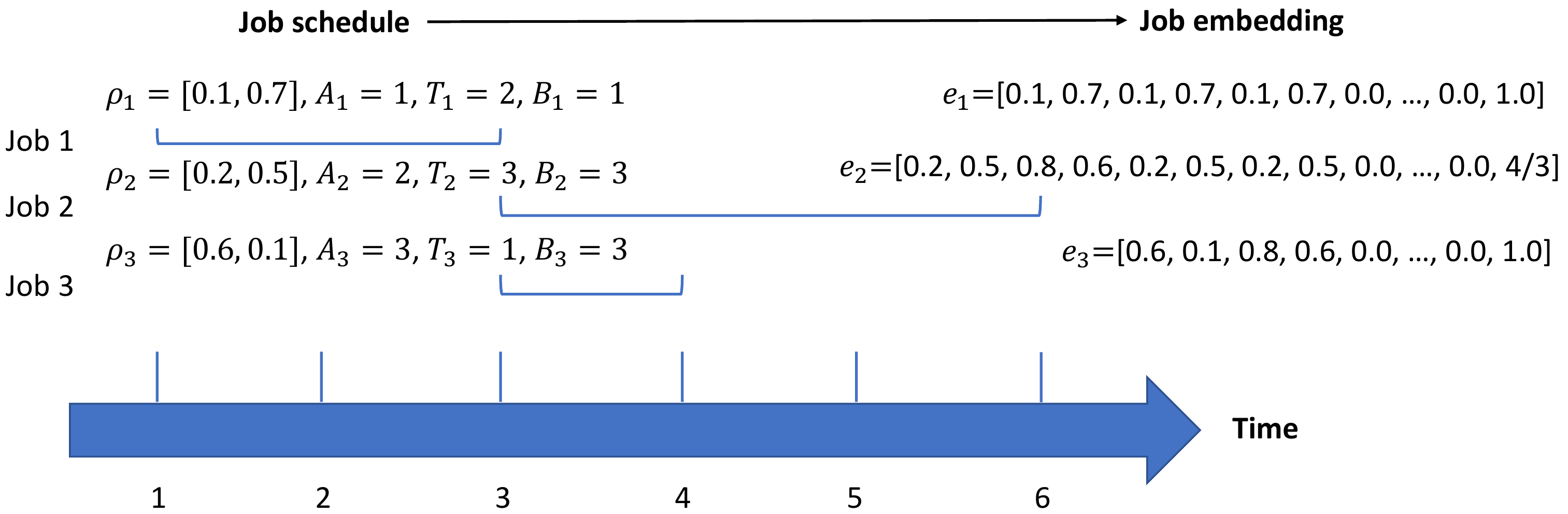}
    \caption{An example to illustrate the job embedding approach for the job scheduling problem.}
    \label{fig:job-embedding}
\end{figure*}

\begin{figure*}[t]
    \centering
    \includegraphics[width=\linewidth]{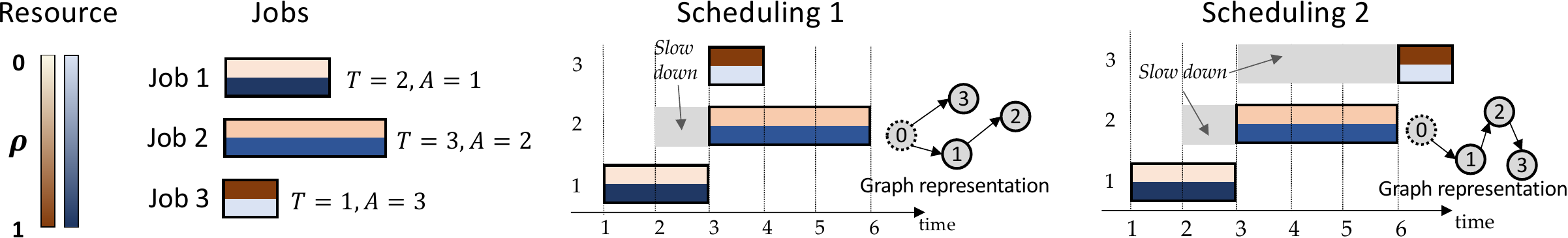}
    \caption{
    An example to illustrate two possible job schedules on a single machine and their corresponding graph representations. Node 0 was added to represent the start of the scheduling process. For multiple machines, multiple node 0 will be added.}
    \label{fig:job-graph}
\end{figure*}

\paragraph{Job embedding.} We embed each job into a $(D \times (T_{max} + 1) + 1)$-dimensional vector $e_j$, where $T_{max}$ is the maximal duration of a job. This vector encodes the information of the job attributes and the machine status during its execution. We describe the details of job embedding as follows. Consider a job $v_j = (\boldsymbol{\rho}_j, A_j, T_j)$. We denote the amount of resources occupied by all jobs at each timestep $t$ as $\boldsymbol{\rho}'_t = (\rho'_{t1}, \rho'_{t2}, ..., \rho'_{tD})$. Each job $v_j$ is represented as a $(D \times (T_{max} + 1) + 1)$-dimensional vector, where the first $D$ dimensions of the vector are $\boldsymbol{\rho}_j$, representing its resource requirement. The following $D \times T_j$ dimensions of the vector are the concatenation of $\boldsymbol{\rho}'_{B_j}, \boldsymbol{\rho}'_{B_j + 1}, ..., \boldsymbol{\rho}'_{B_j + T_j - 1}$, which describes the machine usage during the execution of the job $v_j$. When $T_j < T_{max}$, the following $D \times (T_{max} - T_j)$ dimensions are zero. The last dimension of the embedding vector is the slowdown of the job in the current schedule. We denote the embedding of each job $v_j$ as $e_j$. The embedding of the machine (i.e., $v_0$) is a zero vector $e_0=\mathbf{0}$. Figure~\ref{fig:job-embedding} shows an example of our job embedding approach, and Figure~\ref{fig:job-graph} illustrates an example of the graph construction.

\textbf{Model specification}. To encode the graphs, we extend the Child-Sum Tree-LSTM architecture in~\cite{tai2015improved}, which is similar to the DAG-structured LSTM in~\cite{zhu2016dag}. Specifically, for a job $v_j$, suppose $(h_1, c_1), (h_2, c_2), ..., (h_p, c_p)$ are the LSTM states of all parents of $v_j$, then its LSTM state is:

\begin{equation}
    (h, c) = \mathrm{LSTM} ((\sum_{i=1}^ph_i, \sum_{i=1}^pc_i), e_j)
    \label{eq:jsp-LSTM}
\end{equation}

For each node, the $d$-dimensional hidden state $h$ is used as the embedding for other two components.

\paragraph{Score predictor.} This component is an $L_P$-layer fully connected neural network with a hidden size of $N_P$, and the input to the predictor of job $v_j$ is $h_j$.

\paragraph{Rule selector.} The rewriting rules are equivalent to moving the current job $v_j$ to be a child of another job $v_{j'}$ or $v_0$ in the graph, which means allocating job $v_j$ after job $v_{j'}$ finishes or at its arrival time $A_j$. Thus, the input to the rule selector not only includes $h_j$, but also $h_{j'}$ of all other $v_{j'}$ that could be used for rewriting. The rule selector has two modules. The first module is an $L_S$-layer fully connected neural network with a hidden size of $N_S$. For each job $v_j$, let $N_j$ be the number of jobs that could be the parent of $v_j$, and $\{v_{j'_k}\}$ denotes the set of such jobs. For each $v_{j'_k}$, the input is $[h_j; h_{j'_k}]$, and this module computes a $d$-dimensional vector $h'_k$ to encode such a pair of jobs. The second module of the rule selector is another $L_S$-layer fully connected neural network with a hidden size of $N_S$. For this module, the input is a $(|\mathcal{U}|\times d)$-dimensional vector $[h'_1; h'_2; ...; h'_{|\mathcal{U}|}]$, where $|\mathcal{U}|=2W$. When $N_j < |\mathcal{U}|$, $h'_{N_j + 1}, h'_{N_j + 2}, ..., h'_{|\mathcal{U}|}$ are set to be zero. The output layer of this module is a $|\mathcal{U}|$-dimensional softmax layer, which predicts the probability of each different move of $v_j$.

\subsection{More Details for Vehicle Routing Problem}
\label{app:model-details-vrp}

\paragraph{Node embedding.} We embed each node into a $7$-dimensional vector $e_j$. This vector encodes the information of the node position, node resource demand, and the current status of the vehicle. We describe the details of node embedding as follows. Consider a node $v_j = ((x_j, y_j), \delta_j)$, where $(x_j, y_j)$ is the position, and $\delta_j$ is the resource demand. We set $\delta_0=0$ for the depot (i.e., node 0). Denote $Cap$ as the vehicle capacity. The first three dimensions of $e_j$ are $x_j$, $y_j$, and $\delta_j / Cap$. The next three dimensions of $e_j$ are the coordinates of the node visited at the previous step (set as the depot position for the first visited node) and the Euclidean distance between $v_j$ and the previous node. The last dimension is the amount of remaining resources carried by the vehicle at the current step, which is also normalized by the vehicle capacity.

\paragraph{Score predictor.} This component is an $L_P$-layer fully connected neural network with a hidden size of $N_P$, and the input to the predictor of the node $v_j$ is $h_j$, where $h_j$ is the output of the bi-directional LSTM used to encode each node in the route.

\paragraph{Rule selector.} The rewriting rules are equivalent to moving a node in the route $v_j$ after another node $v_{j'}$, similar to the job scheduling setting. However, different from job scheduling, the number of such nodes $v_{j'}$ varies among different problems. Thus, we train an attention module to select $v_{j'}$, with a similar design to the pointer network~\cite{vinyals2015pointer}.

\subsection{Model hyper-parameters}

For both the expression simplification and job scheduling tasks, $L_S=L_P=1$. For the vehicle routing task, $L_S=L_P=2$. For all the three tasks, $N_S=N_P=256$, $d=512$.

\section{More Details on Training}
\label{app:training-details}

\begin{algorithm*}[t]
\caption{Forward Pass Algorithm for the Neural Rewriter during Training}
\label{alg:fwd}
\begin{algorithmic}[1]
\Require initial state $s_0$, hyper-parameters $\epsilon, p_c, T_{iter}, T_{\omega}, T_{u}$
\For {$t = 0 \to T_{iter} - 1$}
    \For{$i = 1 \to T_{\omega}$}
        \State Sample $\omega_t\sim\pi_\omega(\omega_t|s_t; \theta)$, where $\omega_t \in \Omega(s_t)$, $\pi_\omega(\omega_t|s_t; \theta) = \frac{\exp(Q(s_t, \omega_t; \theta))}{\sum_{\omega_t} \exp(Q(s_t, \omega_t; \theta))}$
        \If {$Q(s_t, \omega_t; \theta) < \epsilon$}
            \State Re-sample $\omega'_t\sim\pi_\omega(\omega'_t|s_t; \theta)$ with a probability of $1-p_c$
            \State \ifs{\text{Re-sampling is not performed}}{\textbf{break}}
        \Else
            \State \textbf{break}
        \EndIf
    \EndFor
    \For {$i=1 \to T_{u}$}
        \State Sample $u_t\sim \pi_u(u_t|s_t[\omega_t]; \phi)$
        \State \ifs{u_t \text{ can be applied to } s_t[\omega_t]}{\textbf{break}}
    \EndFor
    \State \ifs{u_t\text{ does not applied to }s_t[\omega_t]}{\textbf{break}}
    \State $s_{t+1} = f(s_t, \omega_t, u_t)$
\EndFor
\end{algorithmic}
\end{algorithm*}

Algorithm~\ref{alg:fwd} presents the details of the forward pass during training. The forward pass during evaluation is similar, except that we compute $\omega_t$ and $u_t$ as $\omega_t = \arg\max_{\omega}\pi_\omega(\omega|s_t; \theta)$ and $u_t = \arg\max_{u}(\pi_u(u|s_t[\omega_t]; \phi))$, and the inference immediately terminates when $Q(s_t, \omega_t; \theta) < \epsilon$ or $u_t$ does not apply.

\eat{
\paragraph{Reward design.} For job scheduling problem, we simply define the reward function as $r(s_t, (\hat{g}_t, a_t)) = c(s_t) - c(s_{t+1})$. The same reward function does not apply to the expression simplification problem, since it would always give a negative reward for uphill rewriting rules such as expanding a ``min/max''. Thus, for expression simplification problem, we modify the reward function to $r(s_t, (\hat{g}_t, a_t)) = max_{t+1 \leq k \leq T}(\gamma^{k-t}(c(s_t) - c(s_k))) / c(\hat{g}_t)$, where $\gamma$ is a discount factor. This design of reward function would assign a positive value to an uphill rule if it is included in a path that results in a simplified expression in the end. We normalize the reward by $c(\hat{g}_t)$, so that the reward function is bounded. In our evaluation, we set $\gamma=0.9$.}

\paragraph{Hyper-parameters.} For all tasks in our evaluation, in Algorithm~\ref{alg:fwd}, $\epsilon=0.0, T_{\omega}=10, T_{u}=10$. $T_{iter}=50$ for both the expression simplification and the job scheduling tasks. For the vehicle routing task, we set $T_{iter}=200$, because we find that applying 50 rewriting steps could be insufficient for finding a competitive solution, especially when the number of customer nodes is large. For all tasks in our evaluation, $p_c$ is initialized with $0.5$, and is decayed by $0.8$ for every 1000 timesteps until $p_c=0.01$, where it is not decayed anymore. In the training loss function, $\alpha = 10.0$. The decay factor for the cumulative reward is $\gamma=0.9$. The initial learning rate is $1e-4$, and is decayed by $0.9$ for every 1000 timesteps. Batch size is 128. Gradients with $L_2$ norm larger than 5.0 are scaled down to have the norm of 5.0. The model is trained using Adam optimizer. All weights are initialized uniformly randomly in $[-0.1, 0.1]$.

\section{More Results for Job Scheduling Problem}
\label{app:jsp-res}

We observe that while OR-tools is a high-performance solver for generic combinatorial optimization problems, it is less effective than both heuristic-based scheduling algorithms and neural network approaches on our job scheduling problem, especially with more resource types. After looking into the schedules computed by OR-tools, we find that they often prioritize long jobs over short jobs, while swapping the scheduling order between them would clearly decrease the job waiting time. On the other hand, both our neural rewriter and heuristic algorithms based on the job length would usually schedule short jobs very soon after their arrival, which results in better schedules.

Table~\ref{tab:res-jsp-frequency} and~\ref{tab:res-jsp-resource} present the results of ablation study on job frequency and resource distribution respectively.

To examine how the initial schedules affect the final results, besides earliest-job-first schedules, we also evaluate initial schedules with different average slowdown. Specifically, for each job sequence, we generate different initial schedules by randomly allocating one job at a time.

In Table~\ref{tab:jsp-diff-init-schedule}, we present the results with $D=20$ types of resources. For each job sequence, we randomly generate 10 different initial schedules. We can observe that although the effectiveness of initial schedules affects the final schedules, the performance is still consistently better than other baseline approaches, which demonstrates that our neural rewriter is able to substantially improve the initial solution regardless of its quality.

\begin{table*}[t]
    \centering
    \begin{tabular}{|c|c|c|}
    \hline
    & Dynamic Job Frequency & Steady Job Frequency\\
    \hline
    Earliest Job First (EJF) & 14.53 & 24.23 \\
    Shortest Job First (SJF) & 3.62 & 5.00 \\
    SJF-offline & 2.70 & 4.26 \\
    NeuRewriter (dynamic) & \textbf{2.56} & 3.99 \\
    NeuRewriter (steady) & 2.59 & \textbf{3.94} \\
    \hline
    \end{tabular}
    \caption{Experimental results of the job scheduling problem with different distribution of job frequency.}
    \label{tab:res-jsp-frequency}
    \begin{tabular}{|c|c|c|}
    \hline
    & Uniform Job Resources & Non-uniform Job Resources\\
    \hline
    Earliest Job First (EJF) & 11.06 & 24.23 \\
    Shortest Job First (SJF) & 4.51 & 5.00 \\
    SJF-offline & 2.76 & 4.26 \\
    NeuRewriter (uniform) & \textbf{2.73} & 4.05 \\
    NeuRewriter (non-uniform) & 3.13 & \textbf{3.94} \\
    \hline
    \end{tabular}
    \caption{Experimental results of the job scheduling problem with different distribution of job resources.}
    \label{tab:res-jsp-resource}
\end{table*}

\eat{
\begin{table*}[t]
    \centering
    \begin{tabular}{|c|c|c|c|c|}
    \hline
    D  &  2 & 5 & 10 & 20\\
    \hline
    Earliest Job First & 11.11 & 13.62 & 22.13 & 24.23 \\
    Shortest Job First & 4.80 & 5.83 & 5.58 & 5.00 \\
    Shortest First Search & 4.25 & 5.05 & 5.54 & 4.98 \\
    OR-tools & 4.58 & 6.48 & 13.54 & 15.18 \\
    DeepRM~\cite{mao2016resource} & 2.81 & 6.52 & 9.20 & 10.18 \\
    Offline Planning & 2.57 & 3.02 & 4.08 & 4.26\\
    Neural Rewriter (Ours) & \textbf{2.53} & \textbf{2.93} & \textbf{3.81} & \textbf{3.94} \\
    \hline
    \end{tabular}
    \caption{Experimental results of the job scheduling problem on the test set. For each approach, we report the average slowdown of the jobs with different number of resource types $D$.}
    \label{tab:res-jsp}
\end{table*}

\begin{table*}[t]
    \centering
    \begin{tabular}{|c|c|c|c|}
    \hline
    & All Short Jobs & All Long Jobs & Non-uniform Job Lengths\\
    \hline
    Earliest Job First (EJF) & 7.36 & 16.35 & 24.23 \\
    Shortest Job First (SJF) & 3.73 & 14.42 & 5.00 \\
    SJF-Offline & \textbf{3.70} & \textbf{13.80} & 4.26 \\
    Neural Rewriter (trained on jobs of uniform lengths) & 4.34 & 14.17 & 4.56 \\
    Neural Rewriter (trained on jobs of non-uniform lengths) & 4.42 & 15.25 & \textbf{3.94} \\
    \hline
    \end{tabular}
    \caption{Experimental results of the job scheduling problem with different distribution of job lengths.}
    \label{tab:res-jsp-length}
\end{table*}}

\begin{table*}[t]
    \centering
    \begin{tabular}{|c|c|c|c|}
        \hline
         Initial average slowdown & $\leq 10$ & $10-25$ & $>25$ \\
         \hline
         Final average slowdown & 3.88 & 3.90 & 4.06 \\
         \hline
         \hline
         Earliest Job First (EJF) & \multicolumn{3}{|c|}{24.23} \\
         Shortest Job First (SJF) & \multicolumn{3}{|c|}{5.00} \\
         Shortest First Search (SJFS) & \multicolumn{3}{|c|}{4.98} \\
         DeepRM & \multicolumn{3}{|c|}{10.18} \\
         OR-tools & \multicolumn{3}{|c|}{15.18} \\
         SJF-offline & \multicolumn{3}{|c|}{4.26} \\
         NeuRewriter & \multicolumn{3}{|c|}{\textbf{3.94}} \\
         \hline
    \end{tabular}
    \caption{Experimental results of the job scheduling problem using initial schedules with different average slowdown. The number of resource types $D=20$.}
    \label{tab:jsp-diff-init-schedule}
\end{table*}

\section{More Discussion of the Evaluation on Vehicle Routing Problem}
\label{app:vrp-res}

\begin{figure*}
    \begin{subfigure}[t]{0.5\linewidth}
        \centering
        \includegraphics[width=\linewidth]{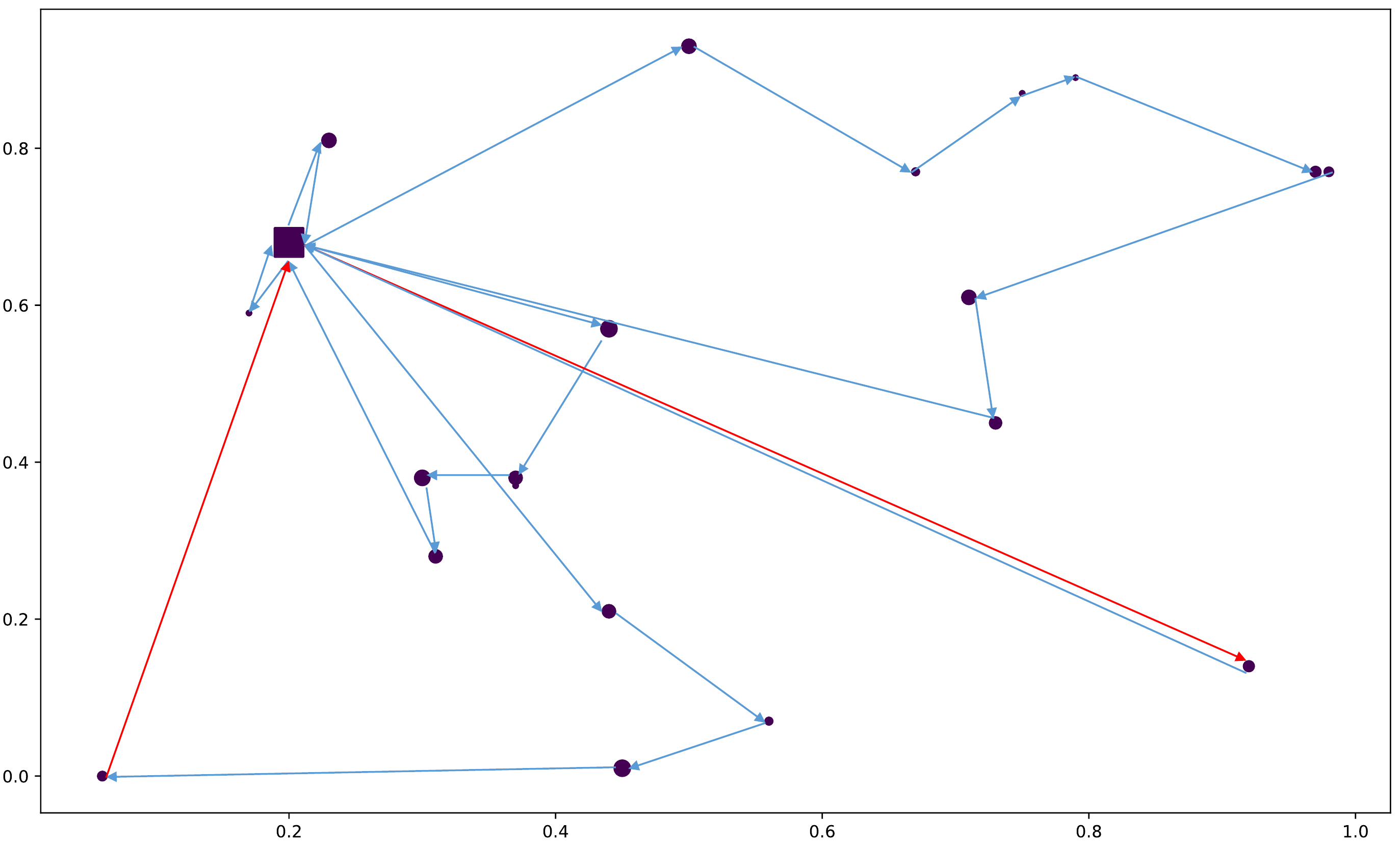}
        \caption{Step 0.}
    \end{subfigure}
    \begin{subfigure}[t]{0.5\linewidth}
        \centering
        \includegraphics[width=\linewidth]{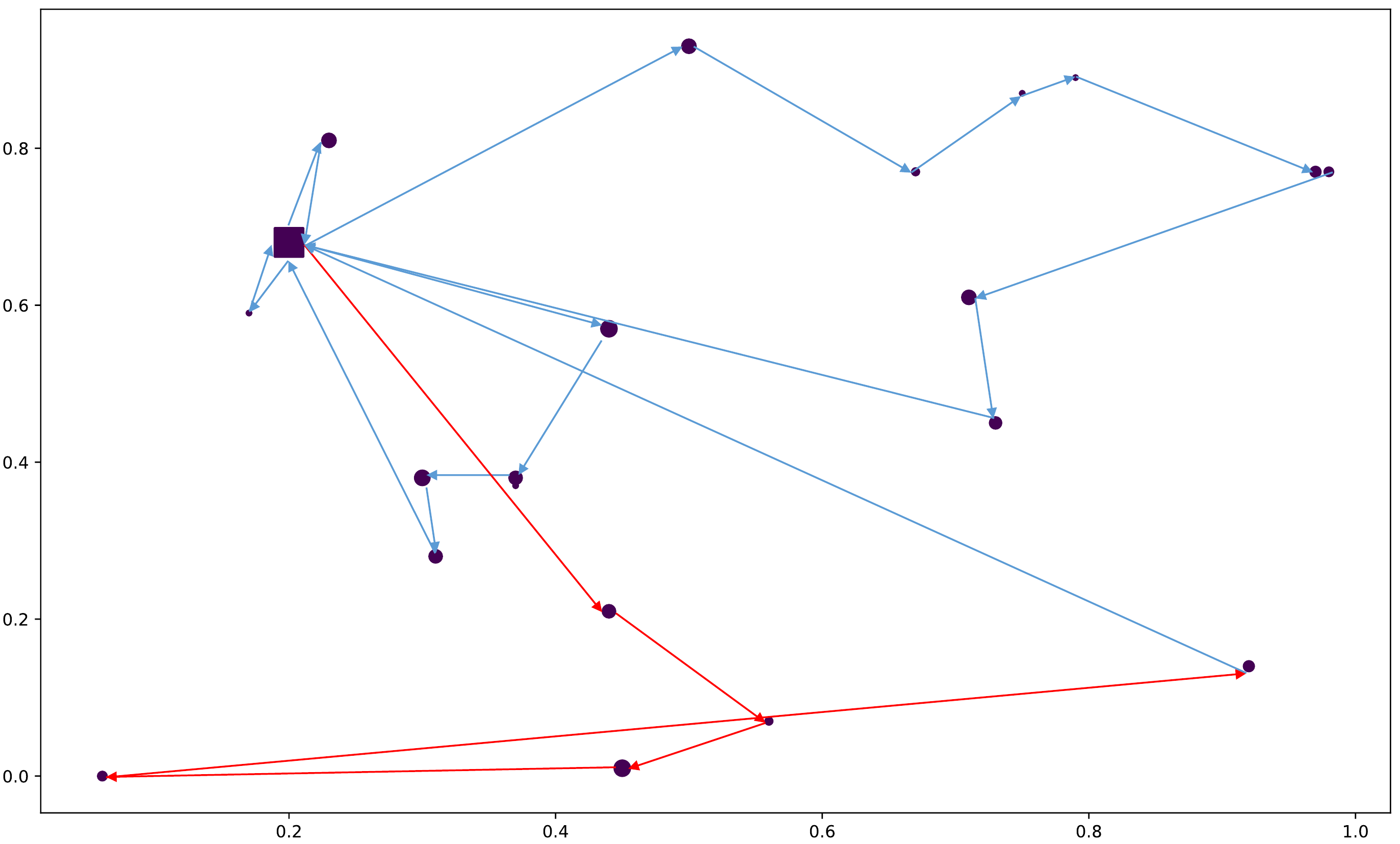}
        \caption{Step 1.}
    \end{subfigure}
    \begin{subfigure}[t]{0.5\linewidth}
        \centering
        \includegraphics[width=\linewidth]{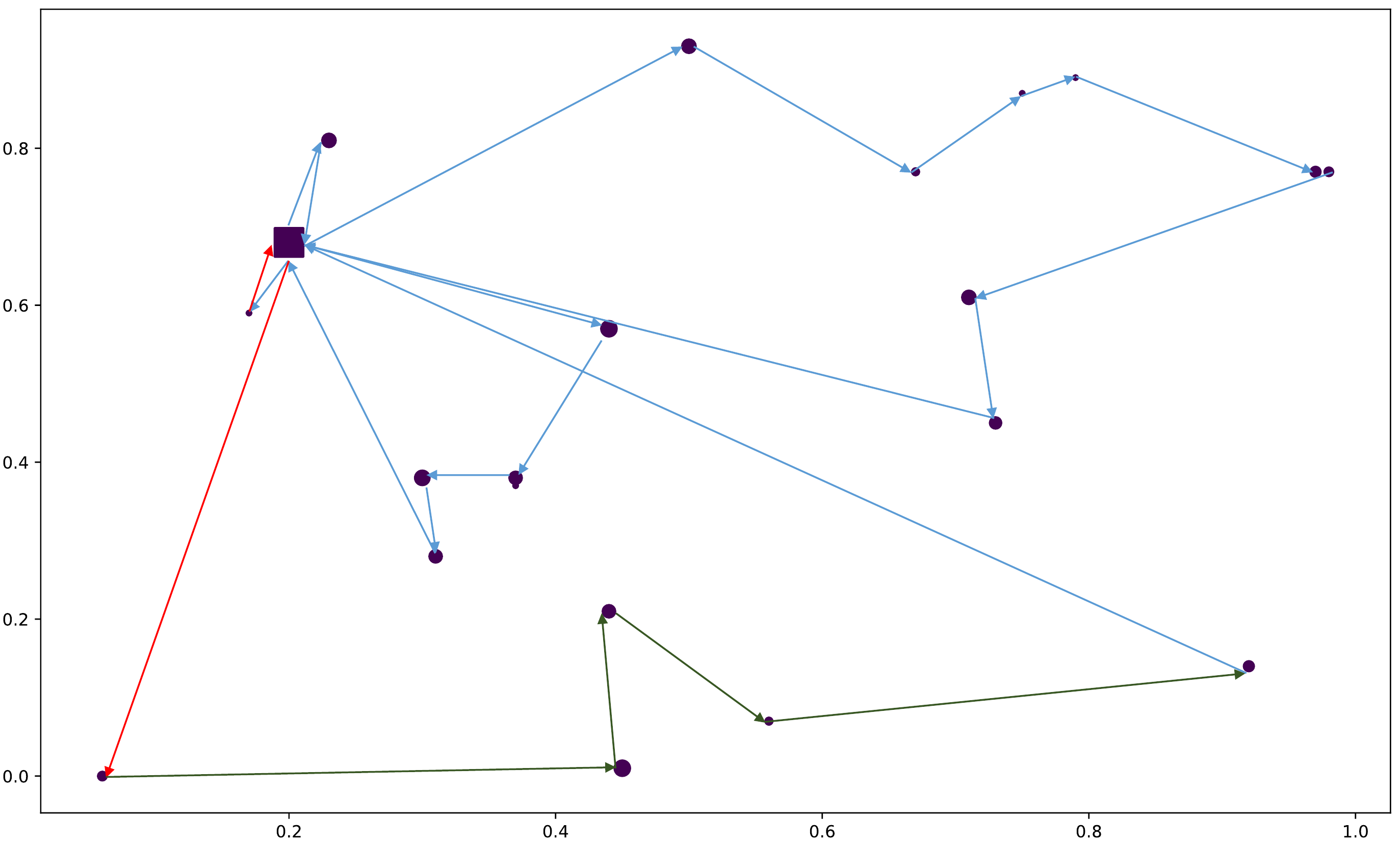}
        \caption{Steps 2-5.}
    \end{subfigure}
    \begin{subfigure}[t]{0.5\linewidth}
        \centering
        \includegraphics[width=\linewidth]{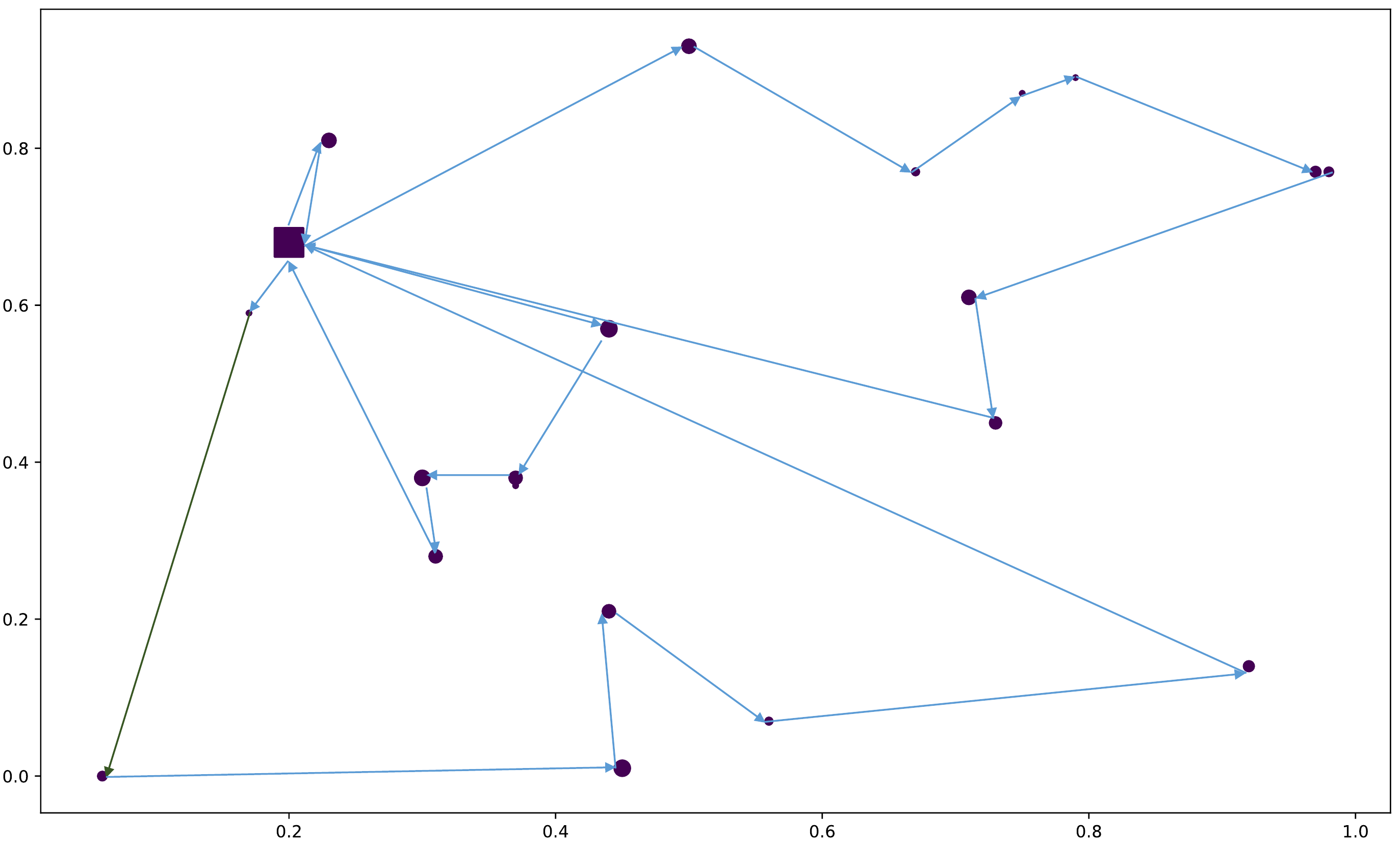}
        \caption{Step 6.}
    \end{subfigure}
    \caption{An example of the rewriting steps for a VRP20 problem. The square is the depot, and circles are customer nodes. The customer node sizes are proportional to their resource demands. At each stage, red edges are to be rewritten at the next step, and green edges are rewritten ones. The tour length of the initial route is 7.31, and the final tour length after rewriting is 5.98.}
    \label{fig:ex-vrp20-0}
\end{figure*}

We generate the initial routes for {\ours} in the following way: starting from the depot, at each timestep, the vehicle visits the nearest node that is either: (1) a customer node that has not been visited yet, and its resource demand can be satisfied; or (2) the depot node, and the resources carried by the vehicle is less than its capacity. See Figure~\ref{fig:ex-vrp20-0} for examples of the initial solutions. In this way, the average tour length is $7.74$ for VRP20, $13.47$ for VRP50, and $20.36$ for VRP100. Note that these results are even worse than the classic heuristics compared in Table~\ref{tab:res-vrp}.

Table~\ref{tab:res-vrp} presents more results for vehicle routing problems, and Figure~\ref{fig:ex-vrp20-0} shows an example of the rewriting steps performed by {\ours}.

For generalization results, note that after training on VRP50, {\ours} achieves an average tour length of 17.33 on VRP100 (See Figure~\ref{fig:res-vrp-gen} in the mainbody of the paper). This is better than 18.00 reported in~\cite{nazari2018reinforcement}, suggesting that our approach could adapt better to different problem distributions.

\begin{table}[t]
    \centering
    \begin{tabular}{|c|c|c|c|}
        \hline
        Model & VRP20, Cap30 & VRP50, Cap40 & VRP100, Cap50 \\
        \hline
        {\ours} & \textbf{6.16} & \textbf{10.51} & \textbf{16.10} \\
        \hline
        AM-Greedy & 6.40 & 10.98 & 16.80 \\
        AM-Sampling & 6.25 & 10.62 & 16.23 \\
        \hline
        Nazari et al. (RL-Greedy) & 6.59 & 11.39 & 17.23 \\
        Nazari et al. (RL-BS(5)) & 6.45 & 11.22 & 17.04 \\
        Nazari et al. (RL-BS(10)) & 6.40 & 11.15 & 16.96 \\
        \hline
        CW-Greedy & 7.22 & 12.85 & 19.72 \\
        CW-Rnd(5,5) & 6.89 & 12.35 & 19.09 \\
        CW-Rnd(10,10) & 6.81 & 12.25 & 18.96 \\
        \hline
        SW-Basic & 7.59 & 13.61 & 21.01 \\
        SW-Rnd(5) & 7.17 & 13.09 & 20.47 \\
        SW-Rnd(10) & 7.08 & 12.96 & 20.33 \\
        \hline
        OR-Tools & 6.43 & 11.31 & 17.16 \\
        \hline
        Gurobi (optimal) & 6.10 & - & - \\
        \hline
    \end{tabular}
    \caption{Experimental results of the vehicle routing problems.}
    \label{tab:res-vrp}
\end{table}

\section{More Results for Expression Simplification}
\label{app:Halide-res}

\begin{figure*}
    \begin{subfigure}[t]{0.5\linewidth}
        \centering
        \includegraphics[width=0.7\linewidth]{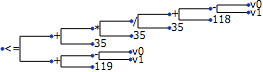}
        \caption{Step 0.}
    \end{subfigure}
    \begin{subfigure}[t]{0.5\linewidth}
        \centering
        \includegraphics[width=0.7\linewidth]{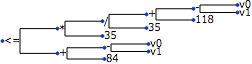}
        \caption{Step 1.}
    \end{subfigure}
    \begin{subfigure}[t]{0.5\linewidth}
        \centering
        \includegraphics[width=0.7\linewidth]{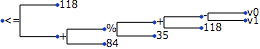}
        \caption{Step 2.}
    \end{subfigure}
    \begin{subfigure}[t]{0.5\linewidth}
        \centering
        \includegraphics[width=0.7\linewidth]{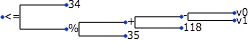}
        \caption{Step 3.}
    \end{subfigure}
    \begin{subfigure}[t]{0.5\linewidth}
        \centering
        \includegraphics[width=0.7\linewidth]{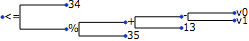}
        \caption{Step 4.}
    \end{subfigure}
    \caption{The rewriting process that simplifies the expression $((v0 - v1 + 18) / 35 * 35 + 35) \leq v0 - v1 + 119$ to $34 \leq (v0 - v1 + 13) \% 35$.}
    \label{fig:ex-Halide-0}
\end{figure*}

\begin{figure*}
        \begin{subfigure}[t]{0.5\linewidth}
            \centering
            \includegraphics[width=0.7\linewidth]{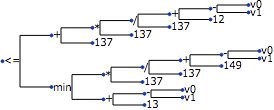}
            \caption{Step 0.}
        \end{subfigure}
        \begin{subfigure}[t]{0.5\linewidth}
            \centering
            \includegraphics[width=0.7\linewidth]{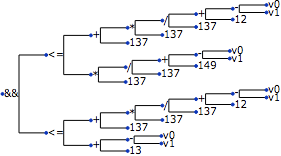}
            \caption{Step 1.}
        \end{subfigure}
        \begin{subfigure}[t]{0.5\linewidth}
            \centering
            \includegraphics[width=0.7\linewidth]{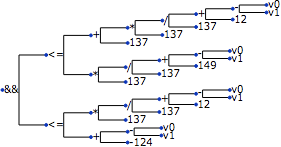}
            \caption{Step 2.}
        \end{subfigure}
        \begin{subfigure}[t]{0.5\linewidth}
            \centering
            \includegraphics[width=0.7\linewidth]{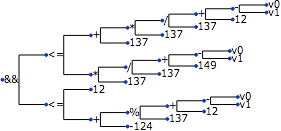}
            \caption{Step 3.}
        \end{subfigure}
        \begin{subfigure}[t]{0.5\linewidth}
            \centering
            \includegraphics[width=0.7\linewidth]{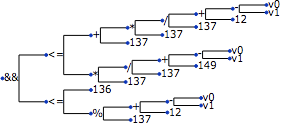}
            \caption{Step 4.}
        \end{subfigure}
            \begin{subfigure}[t]{0.5\linewidth}
            \centering
            \includegraphics[width=0.7\linewidth]{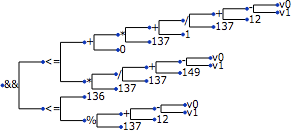}
            \caption{Step 5.}
        \end{subfigure}
        \begin{subfigure}[t]{0.5\linewidth}
            \centering
            \includegraphics[width=0.7\linewidth]{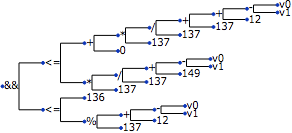}
            \caption{Step 6.}
        \end{subfigure}
        \begin{subfigure}[t]{0.5\linewidth}
            \centering
            \includegraphics[width=0.7\linewidth]{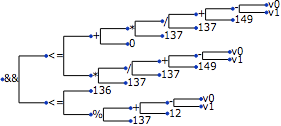}
            \caption{Step 7.}
        \end{subfigure}
        \begin{subfigure}[t]{0.5\linewidth}
            \centering
            \includegraphics[width=0.7\linewidth]{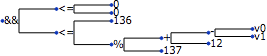}
            \caption{Step 8.}
        \end{subfigure}
        \begin{subfigure}[t]{0.5\linewidth}
            \centering
            \includegraphics[width=0.7\linewidth]{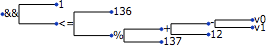}
            \caption{Step 9.}
        \end{subfigure}
        \begin{subfigure}[t]{0.5\linewidth}
            \centering
            \includegraphics[width=0.7\linewidth]{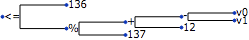}
            \caption{Step 10.}
        \end{subfigure}
    \caption{The rewriting process that simplifies the expression $((v0 - v1 + 12) / 137 * 137 + 137) \leq min((v0 - v1 + 149) / 137 * 137, v0 - v1 + 13)$ to $136 \leq (v0 - v1 + 12) \% 137$.\yuandongg{Correct the subscripts. \% -> mod. Also highlight the changed subtrees.}}
    \label{fig:ex-Halide-1}
\end{figure*}

In Figures~\ref{fig:ex-Halide-0} and~\ref{fig:ex-Halide-1}, we present some success cases of expression simplification, where we can simplify better than both the Halide rule-based rewriter and the Z3 solver.

\eat{
\section{More Discussion on SAT}
\label{app:sat}

We train our neural rewriter on problems with $3-10$ variables, and evaluate on problems with a similar or larger number of variables. We randomly generate $80K$ problems for training, $10K$ for validation, and $10K$ for testing. We set the initial assignment to be $True$ for each variable.

To measure the generalizability of our approach, besides evaluating on problems with a similar number of variables to those in the training set, we also evaluate on problems with a larger number of variables. On problems with 20 variables, our neural rewriter can solve $72\%$ of them, where both DG-DAGRNN and NeuroSAT can solve $69\%$ of them. On problems with 40 variables, our neural rewriter can solve $30\%$ of them. This generalization result is not as good as DG-DAGRNN ($41\%$) and NeuroSAT ($36\%$). We hypothesize that when the number of variables becomes large, it may be insufficient to only flip one variable at each rewriting step. We believe that developing more sophisticated rewriting rules could improve the generalizability of our approach, and we consider it as future work.\yuandong{This sounds quite negative, we might need to change the tone a bit, and analyze possible issue.}
}

%% file: main2.bbl
\begin{thebibliography}{10}

\bibitem{affenzeller2002generic}
M.~Affenzeller and R.~Mayrhofer.
\newblock Generic heuristics for combinatorial optimization problems.
\newblock In {\em Proc. of the 9th International Conference on Operational
  Research}, pages 83--92, 2002.

\bibitem{allamanis2017learning}
M.~Allamanis, P.~Chanthirasegaran, P.~Kohli, and C.~Sutton.
\newblock Learning continuous semantic representations of symbolic expressions.
\newblock In {\em International Conference on Machine Learning}, pages 80--88,
  2017.

\bibitem{armbrust2010view}
M.~Armbrust, A.~Fox, R.~Griffith, A.~D. Joseph, R.~Katz, A.~Konwinski, G.~Lee,
  D.~Patterson, A.~Rabkin, I.~Stoica, et~al.
\newblock A view of cloud computing.
\newblock {\em Communications of the ACM}, 53(4):50--58, 2010.

\bibitem{bachmair1994rewrite}
L.~Bachmair and H.~Ganzinger.
\newblock Rewrite-based equational theorem proving with selection and
  simplification.
\newblock {\em Journal of Logic and Computation}, 4(3):217--247, 1994.

\bibitem{bar1981linear}
R.~Bar-Yehuda and S.~Even.
\newblock A linear-time approximation algorithm for the weighted vertex cover
  problem.
\newblock {\em Journal of Algorithms}, 2(2):198--203, 1981.

\bibitem{bay2017approximating}
A.~Bay and B.~Sengupta.
\newblock Approximating meta-heuristics with homotopic recurrent neural
  networks.
\newblock {\em arXiv preprint arXiv:1709.02194}, 2017.

\bibitem{bello2016neural}
I.~Bello, H.~Pham, Q.~V. Le, M.~Norouzi, and S.~Bengio.
\newblock Neural combinatorial optimization with reinforcement learning.
\newblock {\em arXiv preprint arXiv:1611.09940}, 2016.

\bibitem{blazewicz1996job}
J.~B{\l}a{\.z}ewicz, W.~Domschke, and E.~Pesch.
\newblock The job shop scheduling problem: Conventional and new solution
  techniques.
\newblock {\em European journal of operational research}, 93(1):1--33, 1996.

\bibitem{bradtke1994adaptive}
S.~J. Bradtke, B.~E. Ydstie, and A.~G. Barto.
\newblock Adaptive linear quadratic control using policy iteration.
\newblock In {\em Proceedings of the American control conference}, volume~3,
  pages 3475--3475. Citeseer, 1994.

\bibitem{bunel2016adaptive}
R.~R. Bunel, A.~Desmaison, P.~K. Mudigonda, P.~Kohli, and P.~Torr.
\newblock Adaptive neural compilation.
\newblock In {\em Advances in Neural Information Processing Systems}, pages
  1444--1452, 2016.

\bibitem{cai2018learning}
C.-H. Cai, Y.~Xu, D.~Ke, and K.~Su.
\newblock Learning of human-like algebraic reasoning using deep feedforward
  neural networks.
\newblock {\em Biologically Inspired Cognitive Architectures}, 25:43--50, 2018.

\bibitem{chen2018learning}
T.~Chen, L.~Zheng, E.~Yan, Z.~Jiang, T.~Moreau, L.~Ceze, C.~Guestrin, and
  A.~Krishnamurthy.
\newblock Learning to optimize tensor programs.
\newblock {\em NIPS}, 2018.

\bibitem{chen2017deep}
W.~Chen, Y.~Xu, and X.~Wu.
\newblock Deep reinforcement learning for multi-resource multi-machine job
  scheduling.
\newblock {\em arXiv preprint arXiv:1711.07440}, 2017.

\bibitem{cohn2009sentence}
T.~A. Cohn and M.~Lapata.
\newblock Sentence compression as tree transduction.
\newblock {\em Journal of Artificial Intelligence Research}, 34:637--674, 2009.

\bibitem{de2008z3}
L.~De~Moura and N.~Bj{\o}rner.
\newblock Z3: An efficient smt solver.
\newblock In {\em International conference on Tools and Algorithms for the
  Construction and Analysis of Systems}, pages 337--340. Springer, 2008.

\bibitem{deudon2018learning}
M.~Deudon, P.~Cournut, A.~Lacoste, Y.~Adulyasak, and L.-M. Rousseau.
\newblock Learning heuristics for the tsp by policy gradient.
\newblock In {\em International Conference on the Integration of Constraint
  Programming, Artificial Intelligence, and Operations Research}, pages
  170--181. Springer, 2018.

\bibitem{evans2018can}
R.~Evans, D.~Saxton, D.~Amos, P.~Kohli, and E.~Grefenstette.
\newblock Can neural networks understand logical entailment?
\newblock {\em ICLR}, 2018.

\bibitem{feblowitz2013sentence}
D.~Feblowitz and D.~Kauchak.
\newblock Sentence simplification as tree transduction.
\newblock In {\em Proceedings of the Second Workshop on Predicting and
  Improving Text Readability for Target Reader Populations}, pages 1--10, 2013.

\bibitem{GoogleOrTools}
Google.
\newblock Google or-tools.
\newblock https://developers.google.com/optimization/, 2019.

\bibitem{grandl2015multi}
R.~Grandl, G.~Ananthanarayanan, S.~Kandula, S.~Rao, and A.~Akella.
\newblock Multi-resource packing for cluster schedulers.
\newblock {\em ACM SIGCOMM Computer Communication Review}, 44(4):455--466,
  2015.

\bibitem{graves2014neural}
A.~Graves, G.~Wayne, and I.~Danihelka.
\newblock Neural turing machines.
\newblock {\em arXiv preprint arXiv:1410.5401}, 2014.

\bibitem{haarnoja2017reinforcement}
T.~Haarnoja, H.~Tang, P.~Abbeel, and S.~Levine.
\newblock Reinforcement learning with deep energy-based policies.
\newblock In {\em ICML}, pages 1352--1361. JMLR. org, 2017.

\bibitem{halide2018simplifier}
Halide.
\newblock Halide simplifier.
\newblock https://github.com/halide/Halide, 2018.

\bibitem{hsiang1992term}
J.~Hsiang, H.~Kirchner, P.~Lescanne, and M.~Rusinowitch.
\newblock The term rewriting approach to automated theorem proving.
\newblock {\em The Journal of Logic Programming}, 14(1-2):71--99, 1992.

\bibitem{huang2018gamepad}
D.~Huang, P.~Dhariwal, D.~Song, and I.~Sutskever.
\newblock Gamepad: A learning environment for theorem proving.
\newblock {\em arXiv preprint arXiv:1806.00608}, 2018.

\bibitem{jaderberg2017decoupled}
M.~Jaderberg, W.~M. Czarnecki, S.~Osindero, O.~Vinyals, A.~Graves, D.~Silver,
  and K.~Kavukcuoglu.
\newblock Decoupled neural interfaces using synthetic gradients.
\newblock In {\em Proceedings of the 34th International Conference on Machine
  Learning-Volume 70}, pages 1627--1635. JMLR. org, 2017.

\bibitem{karp1972reducibility}
R.~M. Karp.
\newblock Reducibility among combinatorial problems.
\newblock In {\em Complexity of computer computations}, pages 85--103.
  Springer, 1972.

\bibitem{khalil2017learning}
E.~Khalil, H.~Dai, Y.~Zhang, B.~Dilkina, and L.~Song.
\newblock Learning combinatorial optimization algorithms over graphs.
\newblock In {\em Advances in Neural Information Processing Systems}, pages
  6348--6358, 2017.

\bibitem{kool2018attention}
W.~Kool, H.~van Hoof, and M.~Welling.
\newblock Attention, learn to solve routing problems!
\newblock In {\em International Conference on Learning Representations}, 2019.

\bibitem{lederman2018learning}
G.~Lederman, M.~N. Rabe, and S.~A. Seshia.
\newblock Learning heuristics for automated reasoning through deep
  reinforcement learning.
\newblock {\em arXiv preprint arXiv:1807.08058}, 2018.

\bibitem{levine2014learning}
S.~Levine and P.~Abbeel.
\newblock Learning neural network policies with guided policy search under
  unknown dynamics.
\newblock In {\em Advances in Neural Information Processing Systems}, pages
  1071--1079, 2014.

\bibitem{levine2013guided}
S.~Levine and V.~Koltun.
\newblock Guided policy search.
\newblock In {\em International Conference on Machine Learning}, pages 1--9,
  2013.

\bibitem{mao2016resource}
H.~Mao, M.~Alizadeh, I.~Menache, and S.~Kandula.
\newblock Resource management with deep reinforcement learning.
\newblock In {\em Proceedings of the 15th ACM Workshop on Hot Topics in
  Networks}, pages 50--56. ACM, 2016.

\bibitem{mayne1973differential}
D.~Q. MAYNE.
\newblock Differential dynamic programming--a unified approach to the
  optimization of dynamic systems.
\newblock In {\em Control and Dynamic Systems}, volume~10, pages 179--254.
  Elsevier, 1973.

\bibitem{nazari2018reinforcement}
M.~Nazari, A.~Oroojlooy, L.~Snyder, and M.~Takac.
\newblock Reinforcement learning for solving the vehicle routing problem.
\newblock In {\em Advances in Neural Information Processing Systems}, pages
  9861--9871, 2018.

\bibitem{OpenAIdota}
OpenAI.
\newblock Openai dota 2 bot.
\newblock https://openai.com/the-international/, 2018.

\bibitem{paetzold2013text}
G.~H. Paetzold and L.~Specia.
\newblock Text simplification as tree transduction.
\newblock In {\em Proceedings of the 9th Brazilian Symposium in Information and
  Human Language Technology}, 2013.

\bibitem{paszke2017automatic}
A.~Paszke, S.~Gross, S.~Chintala, G.~Chanan, E.~Yang, Z.~DeVito, Z.~Lin,
  A.~Desmaison, L.~Antiga, and A.~Lerer.
\newblock Automatic differentiation in pytorch.
\newblock In {\em NIPS-W}, 2017.

\bibitem{ragan2013halide}
J.~Ragan-Kelley, C.~Barnes, A.~Adams, S.~Paris, F.~Durand, and S.~Amarasinghe.
\newblock Halide: a language and compiler for optimizing parallelism, locality,
  and recomputation in image processing pipelines.
\newblock {\em ACM SIGPLAN Notices}, 48(6):519--530, 2013.

\bibitem{reeves1995modern}
C.~R. Reeves.
\newblock {\em Modern heuristic techniques for combinatorial problems. Advanced
  topics in computer science}, volume~15.
\newblock Mc Graw-Hill, 1995.

\bibitem{schkufza2013stochastic}
E.~Schkufza, R.~Sharma, and A.~Aiken.
\newblock Stochastic superoptimization.
\newblock In {\em ACM SIGARCH Computer Architecture News}, volume~41, pages
  305--316. ACM, 2013.

\bibitem{scully2017optimally}
Z.~Scully, G.~Blelloch, M.~Harchol-Balter, and A.~Scheller-Wolf.
\newblock Optimally scheduling jobs with multiple tasks.
\newblock {\em ACM SIGMETRICS Performance Evaluation Review}, 45(2):36--38,
  2017.

\bibitem{silver2017mastering}
D.~Silver, J.~Schrittwieser, K.~Simonyan, I.~Antonoglou, A.~Huang, A.~Guez,
  T.~Hubert, L.~Baker, M.~Lai, A.~Bolton, et~al.
\newblock Mastering the game of go without human knowledge.
\newblock {\em Nature}, 550(7676):354, 2017.

\bibitem{sorensson2005minisat}
N.~Sorensson and N.~Een.
\newblock Minisat v1. 13-a sat solver with conflict-clause minimization.
\newblock {\em SAT}, 2005(53):1--2, 2005.

\bibitem{sutton1998reinforcement}
R.~S. Sutton, A.~G. Barto, et~al.
\newblock {\em Reinforcement learning: An introduction}.
\newblock 1998.

\bibitem{tai2015improved}
K.~S. Tai, R.~Socher, and C.~D. Manning.
\newblock Improved semantic representations from tree-structured long
  short-term memory networks.
\newblock In {\em Proceedings of the Annual Meeting of the Association for
  Computational Linguistics}, 2015.

\bibitem{tassa2012synthesis}
Y.~Tassa, T.~Erez, and E.~Todorov.
\newblock Synthesis and stabilization of complex behaviors through online
  trajectory optimization.
\newblock In {\em Intelligent Robots and Systems (IROS), 2012 IEEE/RSJ
  International Conference on}, pages 4906--4913. IEEE, 2012.

\bibitem{terekhov2014queueing}
D.~Terekhov, D.~G. Down, and J.~C. Beck.
\newblock Queueing-theoretic approaches for dynamic scheduling: a survey.
\newblock {\em Surveys in Operations Research and Management Science},
  19(2):105--129, 2014.

\bibitem{tian2013hierarchical}
Y.~Tian and S.~G. Narasimhan.
\newblock Hierarchical data-driven descent for efficient optimal deformation
  estimation.
\newblock In {\em Proceedings of the IEEE International Conference on Computer
  Vision}, pages 2288--2295, 2013.

\bibitem{vinyals2015pointer}
O.~Vinyals, M.~Fortunato, and N.~Jaitly.
\newblock Pointer networks.
\newblock In {\em Advances in Neural Information Processing Systems}, pages
  2692--2700, 2015.

\bibitem{vrabie2009adaptive}
D.~Vrabie, O.~Pastravanu, M.~Abu-Khalaf, and F.~L. Lewis.
\newblock Adaptive optimal control for continuous-time linear systems based on
  policy iteration.
\newblock {\em Automatica}, 45(2):477--484, 2009.

\bibitem{zaremba2014learning}
W.~Zaremba, K.~Kurach, and R.~Fergus.
\newblock Learning to discover efficient mathematical identities.
\newblock In {\em Advances in Neural Information Processing Systems}, pages
  1278--1286, 2014.

\bibitem{zhu2016dag}
X.~Zhu, P.~Sobhani, and H.~Guo.
\newblock Dag-structured long short-term memory for semantic compositionality.
\newblock In {\em Proceedings of the 2016 Conference of the North American
  Chapter of the Association for Computational Linguistics: Human Language
  Technologies}, pages 917--926, 2016.

\end{thebibliography}
